\documentclass{article}

\PassOptionsToPackage{numbers, compress}{natbib}

\usepackage[final]{neurips_2021}

\usepackage[utf8]{inputenc} 
\usepackage[T1]{fontenc}    
\usepackage{xcolor}         
\usepackage[colorlinks]{hyperref}      
\usepackage{url}            
\usepackage{booktabs}       
\usepackage{amsfonts}       
\usepackage{amsmath}
\usepackage{mathtools}      %
\usepackage{nicefrac}       
\usepackage{microtype}      
\usepackage[pdftex]{graphicx}
\usepackage{caption}
\usepackage{subcaption}
\usepackage{wrapfig}
\usepackage{multirow}
\usepackage{verbatim}
\usepackage{algorithm}
\usepackage{enumitem}
\usepackage{comment}
\usepackage{algorithmicx}
\usepackage{algorithm}
\usepackage[noend]{algpseudocode}
\algnewcommand{\LineComment}[1]{\State \(\triangleright\) #1} 
\setlist{noitemsep} 
\setlist[itemize]{leftmargin=*}
\usepackage[export]{adjustbox}
\makeatletter
\g@addto@macro{\UrlBreaks}{\UrlOrds}
\makeatother

\definecolor{light-gray}{gray}{0.9}
\newcommand{\code}[1]{\colorbox{light-gray}{\texttt{#1}}}
\newcommand{\blockcomment}[1]{}
\makeatletter
\newcommand\footnoteref[1]{\protected@xdef\@thefnmark{\ref{#1}}\@footnotemark}
\newcommand*{\rom}[1]{\expandafter\@slowromancap\romannumeral #1@}
\makeatother
\DeclareMathAlphabet\mathbfcal{OMS}{cmsy}{b}{n}
\def\et#1#2#3#4{$e_{#1}^{#4} = (p_{#2}, #3, t)$}
\def\etb#1#2#3#4{$\mathbf{e_{#1}^{#4}} = [(p_{#2}, #3, t)]$}
\def\statemb#1#2{$\mathbfcal{H}\mathbf{(e_{#1}^{#2}})$}
\def\state#1#2{$\mathbfcal{X}\mathbf{(e_{#1}^{#2}})$}

\title{Learning to Combine Per-Example Solutions for Neural Program Synthesis}

\author{%
  Disha Shrivastava \thanks{Correspondence to: <dishu.905@gmail.com>}\\
  Mila, Universit\'e de Montr\'eal \\
  Google Research
  \And
  Hugo Larochelle\\
  Mila, Universit\'e de Montr\'eal \\
  Google Research\\
  CIFAR Fellow
  \And 
  Daniel Tarlow\\
   Mila, McGill University\\
Google Research
}

\begin{document}

\maketitle

\begin{abstract}
The goal of program synthesis from examples is to find a computer program that is consistent with a given set of input-output examples. Most learning-based approaches try to find a program that satisfies all examples at once. Our work, by contrast, considers an approach that breaks the problem into two stages: (a) find programs that satisfy only one example, and (b) leverage these {\it per-example solutions} to yield a program that satisfies all examples. We introduce the \emph{Cross Aggregator} neural network module based on a multi-head attention mechanism that learns to combine the cues present in these per-example solutions to synthesize a global solution. Evaluation across programs of different lengths and under two different experimental settings reveal that when given the same time budget, our technique significantly improves the success rate over PCCoder~\cite{zohar2018automatic} and other ablation baselines. The code, data and trained models for our work can be found at: \url{https://github.com/shrivastavadisha/N-PEPS}.
\end{abstract}

\section{Introduction}\label{intro}

Program synthesis from examples tackles the problem of coming up with a computer program that satisfies a given set of Input-Output (IO) examples. Since the space of possible programs is large, an exhaustive search can be extremely time-consuming. Therefore, development of systems for program synthesis that can come up with a solution (program satisfying the given IO examples) within a \emph{limited time}, such that it is practical for real-world applications, is a challenging task.

Neural-guided program synthesis systems~\cite{balog2016deepcoder, zohar2018automatic} try to expedite the search by using a neural network conditioned on the IO examples as a learned heuristic for the search procedure. In these systems, a neural network outputs probabilities over programs or properties of programs (e.g. functions). These probabilities are then utilized to guide a search like depth-first or beam search. These systems try to find a program that satisfies all IO examples \emph{simultaneously}, which under most of the settings can be hard. What if instead, we try to find this program in parts? To understand this motivation, imagine a process wherein a programmer is asked to write a program that satisfies a set of unit test cases. They may begin by figuring out a program that satisfies a subset of unit test cases first, and later modifying the program to incorporate other corner cases. \citet{shi2019frangel} uses this intuition to iteratively refine a program by mining fragments of Java code from partial solutions, based on a set of rules and predefined heuristics. \citet{sed} also uses the same intuition, but in a different application for program repair. 

\begin{figure}[t]
  \centering
  \includegraphics[width=1.0\textwidth]{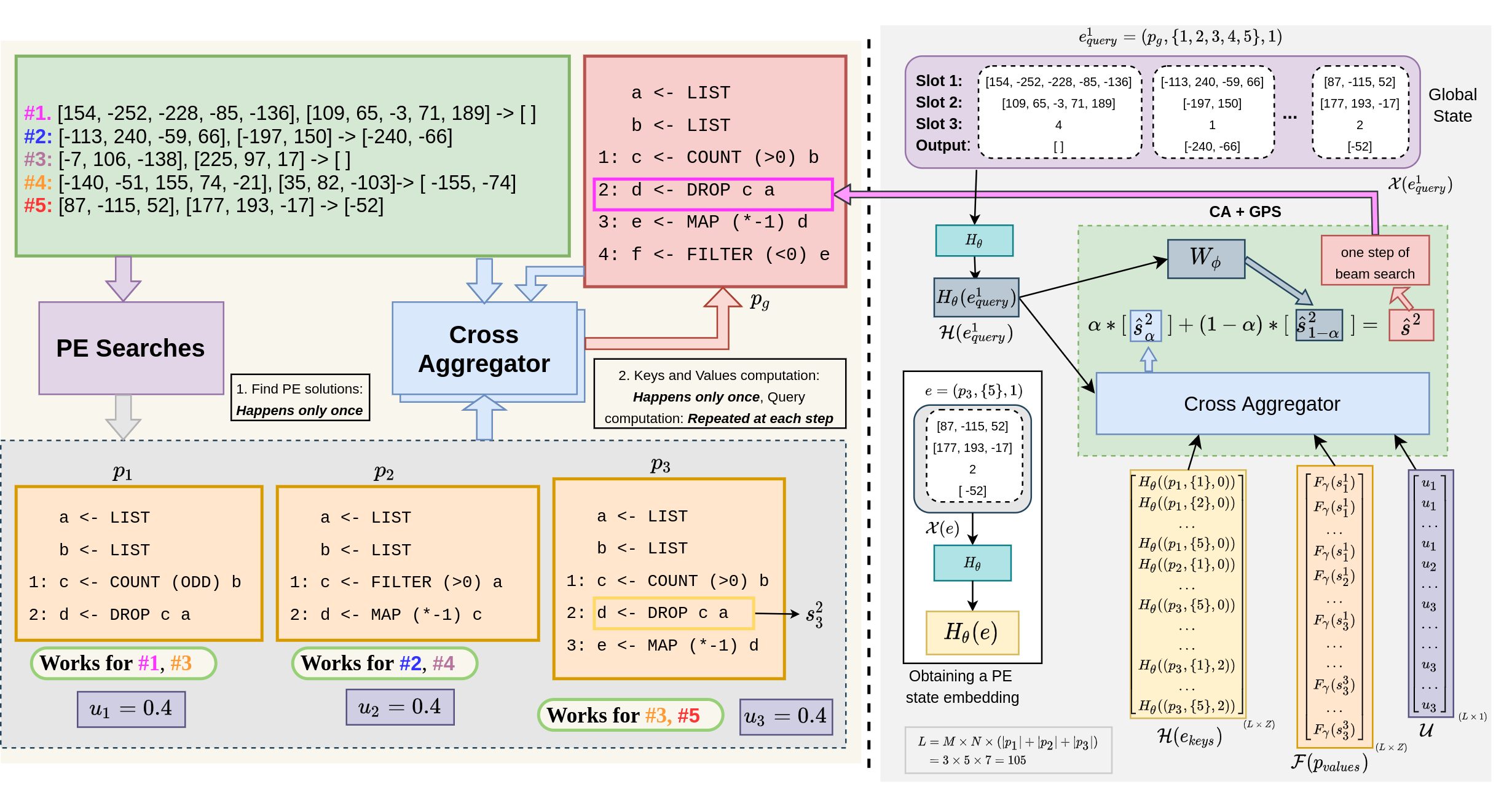}
  \caption{\textbf{Idea of N-PEPS: \emph{(Left)}} Illustrating the two stages of N-PEPS with an example; \textbf{\emph{(Right)}} Synthesizing line 2 of $p_g$ using contributions from CA and GPS, with details of how query, keys, values and relation scores are formed. White box shows an example of obtaining a PE state embedding.}
  \label{fig:block_diagram}
  \vskip -0.15in
\end{figure}

In this work, we consider breaking the complex problem of finding a program that satisfies all $N$ given IO examples (called the \emph{global solution}) into $N$ smaller, easy to solve sub-problems, where each sub-problem involves finding a program satisfying only one IO example (called \emph{per-example solution}). The cues present in these per-example (PE) solutions are then combined to provide useful signals that can help guide the search for the global solution effectively. As a motivating example, consider the left part of Figure~\ref{fig:block_diagram}, where five IO examples are given as a specification (green box) and we need to find a global solution $p_g$ (red box) that satisfies these five examples. The first stage of our approach consists of performing per-example searches to find a program $p_i$ conditioned on the $i$-th IO example. In our example, we start from IO example \#1 and find program $p_1$. In addition, we also check if $p_1$ satisfies any other examples (\#3 in figure). Iterating through the examples in this way results in a set of programs ($p_1, p_2, p_3$) that, taken together, in the ideal scenario, would satisfy all five IO examples. Looking closely at the discovered PE solutions, we see that they contain fragments of the global solution. This brings us to the second stage of our approach that addresses the challenge of how best to aggregate these PE solutions to produce a global solution. Towards that goal, we propose a neural network based architecture, which we refer to as \emph{Cross Aggregator} (CA). It is designed to learn to combine the cues present in these PE solutions, in a way that helps guide the search for $p_g$. We model this aggregation using a multi-head cross-attention mechanism, which leverages the state of step-wise execution of the PE solutions and the synthesized global solution so far (see Section~\ref{sec: ca} for details). Our key contributions can be listed as follows:
\begin{itemize}
    \item We consider breaking the standard program synthesis pipeline into two stages: (a) discovering PE solutions, and (b) aggregating the PE solutions such that it leads to a global solution. We refer to our approach that uses neural networks at both these stages as \emph{Neural Per-Example Program Synthesis }(N-PEPS). 
    \item We propose a neural network based multi-head attention architecture called Cross Aggregator (CA) that makes use of step-wise execution information to \emph{learn} to combine the PE cues such that it helps guide the search for the global solution.
    \item We demonstrate via experiments with programs of different lengths and under two different evaluation settings that when given the same time budget, our formulation shows significant improvements in success rate when compared to PCCoder~\cite{zohar2018automatic} (one of the leading techniques for neural-guided program synthesis) and other ablation baselines. 
\end{itemize}

\section{Background}\label{background}

Suppose we are given a set $X = \{(x_i, y_i)\}_{i=1}^{N}=\{r_i\}_{i=1}^{N}$ of $N$ IO examples and our task is to come up with a program $p_g$ that satisfies these examples. The $i$-th IO example $r_i$ consists of a pair of input $x_i$ and output $y_i$. The program consists of $T$ lines (excluding lines with input variable declarations), i.e.\ $p_g = [p_g^t]_{t=1}^{T}$. To be practically meaningful, we impose the constraint that $p_g$ has to be found within a given time budget, specified by a \emph{timeout} value. The syntax and semantics of $p_g$ are governed by a domain-specific language (DSL). We use the DSL provided by \citet{balog2016deepcoder}, which contains first-order functions (e.g.\ \code{SORT, REVERSE}) and higher-order functions (e.g.\ \code{MAP, FILTER}) that can take \emph{lambda} functions (e.g.~\ (*4), (<0)) as input. The inputs and outputs can be either an integer or a list of integers (see Appendix F of \citet{balog2016deepcoder} for more details about the DSL). The Predict and Collect Coder (PCCoder) \cite{zohar2018automatic} provides state-of-art results for this DSL and is illustrative of methods that directly solve for all available IO examples at once. We refer to these methods as Global Program Search (GPS). We will be building on PCCoder to propose our per-example approach.
\textbf{\begin{figure}[t]
  \centering
  \includegraphics[width=1.0\textwidth]{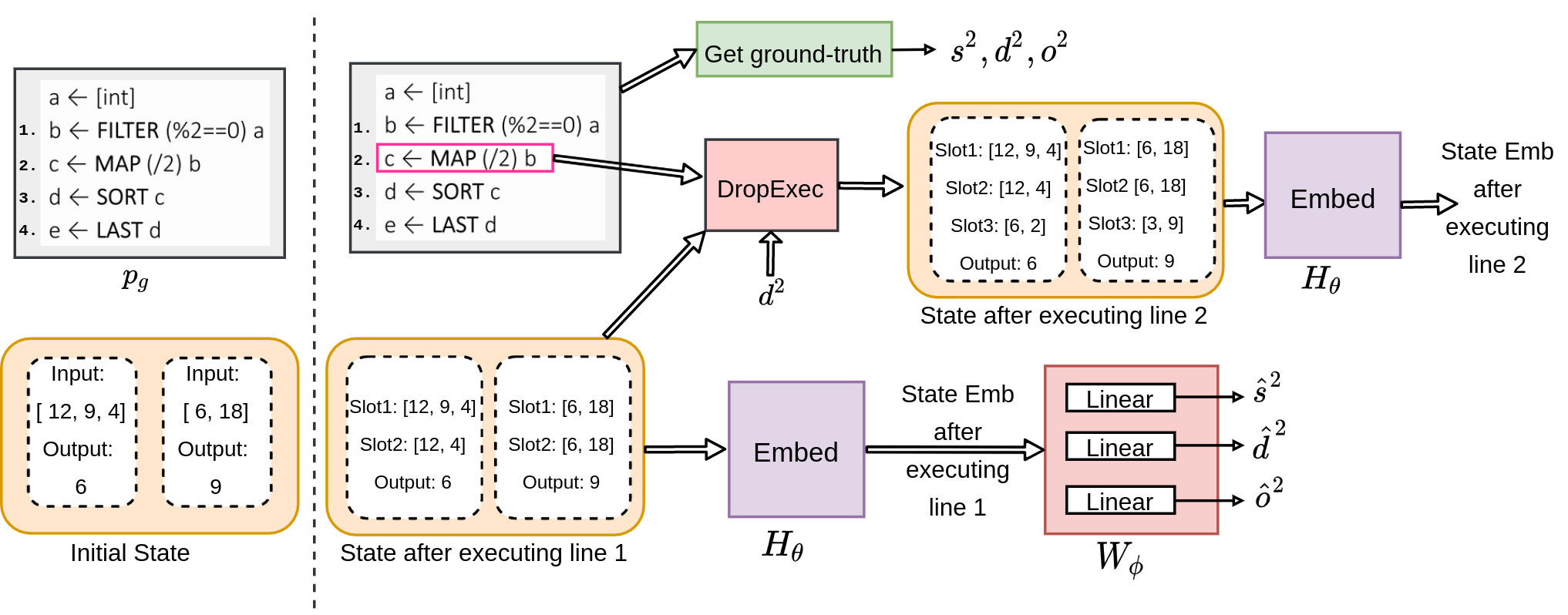}
  \caption{\textbf{\textit{(Left):}} Sample program along with two IO examples that forms the program state at $t=0$; \textbf{\textit{(Right):}} Block Diagram explaining the training of PCCoder at line 2 of the program.}
  \label{fig:example}
  \vskip -0.15in
\end{figure}
}
\subsection{PCCoder} \label{sec:pccoder}

PCCoder synthesizes programs one line at a time, through a model based on the notion of a {\it program state}. The program state is a two-dimensional memory of size $N \times (\nu + 1)$ obtained during the execution of $t$ lines (steps) of a program on a set of $N$ inputs. This means that for each IO example $r_i$, there are up to $\nu$ slots for storing the input and intermediate program variables, with an additional slot for storing the output (see Appendix~\ref{app:state} for more details). Note that the initial state at $t=0$ consists of only the IO examples (see left part of Figure~\ref{fig:example}).

PCCoder consists of two learnable components (i.e.\ neural networks), $H_{\theta}$ and $W_{\phi}$, with parameters $\theta$ and $\phi$. $H_{\theta}$ obtains the embedding of the current program state by average-pooling the representation of the $\nu + 1$ slots corresponding to individual examples (white boxes inside the state in Figure~\ref{fig:example}) into a vector of fixed size in $\mathbb{R}^Z$, where $Z$ denotes the embedding size (see Appendix~\ref{app:state} for details of how these representations of slots are obtained). $W_{\phi}$ maps this \emph{state embedding} to predictions of three quantities of interest for the next line in the program: (a) the next operator $\hat{o}^t$ (or function e.g. \code{MAP}); (b) the next statement $\hat{s}^t$ (operator along with its arguments e.g. \code{MAP(/2) b}); and (c) next drop vector $\hat{d}^t$ which represents positions of variables that can be dropped from the state. The dropping is desirable as it creates slots for storing new variables, which in turn allows for synthesizing longer programs. There is a module called $DropExec$ which executes a given line of the program against an example $r_i$ and stores the resulting variable $c_i$ in the next available slot in the state. If all $\nu$ slots in the state are filled, a variable is dropped from one of the slots using the drop vector and $c_i$ is stored there. The updated state can then be used for predicting the next line (see right part of Figure~\ref{fig:example}). Next, we provide details of how training and inference is done in PCCoder.

\textbf{Training:} For training $H_{\theta}$ and $W_{\phi}$, several instances of a specification $X$ and the ground-truth program $p_g$ are provided. Given an instance and line $t$ of the program, training operates by obtaining the ground-truth values of statements ($s^t$), operator ($o^t$) and drop vector ($d^t$). The statement and operator values are represented as one-hot vectors of size equal to the number of statements ($n_s$) and number of operators ($n_o$), respectively in the DSL. The drop vector is a multi-hot vector of size $\nu$ with ones at positions corresponding to the variables in the program that can be dropped, i.e.\ variables that don't appear in subsequent lines in the program. The step-wise loss $\mathcal{L}$ is the sum of cross-entropy losses between the actual and predicted statement and operator, and the binary cross entropy loss between each position in the actual and predicted drop vector. The task of predicting the operator is an auxiliary task, i.e.\ it is used only during training and not at inference time, and is found to improve the training performance. During training, to obtain the updated state, the $DropExec$ module chooses the drop-index to be a random entry from those positions in the drop vector $d^t$ that are ones. The right part of Figure~\ref{fig:example} illustrates the process of training at step 2.

\textbf{Inference:} Inference is done using complete anytime beam search (CAB)~\citep{cab} where the time for search is upper bounded by the timeout value. The CAB algorithm operates by performing different beam searches repeatedly in an outer loop. The pruning conditions of the beam search (i.e., beam size, expansion size) are weakened with each iteration of the outer loop, until a solution is found. The inner loop consists of different steps of a single beam search. At each step, the beam consists of the most promising program prefixes, with each prefix represented as a tuple of the current program state, synthesized program until now and the product of the probabilities of the statements in the synthesized program. To synthesize the next line of the program, prefixes are expanded by executing the statements in decreasing order of statement probabilities and taking the argmax of the drop vector probabilities. The statement and drop vector probabilities are obtained using the trained neural networks $H_{\theta}$ and $W_{\phi}$. The search terminates if we find a candidate program prefix that satisfies all $N$ IO examples. The corresponding program is the synthesized global solution $p_g$. Note that the search may fail and not discover a global solution within the specified timeout. Appendix~\ref{app:pccoder} gives details of training and inference procedures, and modules of PCCoder.

\section{Neural Per-Example Program Synthesis (N-PEPS)}\label{methodology}
As stated in Section~\ref{intro}, in this work, we decide to break the complex problem of finding a global solution $p_g$ that satisfies all $N$ IO examples, into $N$ smaller sub-problems. Each sub-problem aims to find a program $p_i$ that will satisfy only the IO example $r_i$. The cues present in these PE solutions are then aggregated to help guide the search for $p_g$. We constrain the process of breaking and combining to fit within the specified timeout value. The distribution of total timeout between these stages is treated as a hyperparameter. In this section, we discuss our process of finding PE solutions and follow it with a description of our neural network module that learns to combine the PE solutions. 

\subsection{Per Example Program Synthesis}
\label{sec:peps}
We refer to the general framework of finding PE solutions first and later aggregating the PE cues to find a global solution, as \emph{Per-Example Program Synthesis} (PEPS). We call the module that finds PE solutions as the \emph{PE Searches} module. To train the PE Searches module, we use the PCCoder model as it is, except that it is trained to take a single IO example as input as opposed to all the examples in $X$.  We will call this trained model as the \emph{PE model}. We allocate a fixed value of \emph{PEPS timeout}, which is the maximum time given to find each PE solution. The sum of PEPS timeouts across all PE solutions should be less than the total timeout, so that there is some time left for the CA module to aggregate the PE cues (i.e., $N \times$ PEPS Timeout < Total Timeout). We start from the first example, and using the PE model, try to find a solution that satisfies it. Once found, we also check if this solution satisfies other examples in $X$. We record the fraction of IO examples satisfied by $p_i$, and call it the \emph{PE solution score} $u_i$. If $p_i$ satisfies all examples in $X$ (i.e. $u_i = 1.0$), we stop and return $p_g = p_i$ as the global solution. Otherwise, we proceed to find the next PE solution (based on the order of examples given in $X$). Note that it is possible that for certain examples in $X$, we fail to find a PE solution within the PEPS timeout. Once we have our list of $M$ PE solutions ($0 \leq M \leq N$), which ideally satisfies all $N$ examples but may not necessarily, we proceed to aggregating them. Note that when comparison with baselines is not a requirement, we can increase speedup by finding PE solutions in parallel (see Appendix D.1 for more details).

\subsection{Cross Aggregator}
\label{sec: ca}
\textbf{Notation:} To formulate the program state, we define a basic unit called an \emph{execution tuple} (ET). An ET \et{}{}{\mathcal{S}}{} is a tuple consisting of a program $p$, a subset $\mathcal{S}$ of example indices in $X$ and a step number $t$. Executing the first $t$ steps (lines) of a program $p$ on every example $r_i$ for $i \in \mathcal{S}$ yields a program state which we note as $\mathbfcal{X}(e)$. Like PCCoder, we pool the representation of slots of the state corresponding to each example $r_i$ for $i \in \mathcal{S}$ to obtain a state embedding (see Section~\ref{sec:pccoder}), hence making its size independent of the size of $\mathcal{S}$. To represent different combinations of programs executed against different sets of examples at different time steps, we define a list $\mathbf{e}$ of such execution tuples, with its size denoted by $L$. $(p_1, \{1\}, 0)$ and $(p_3, \{2\}, 2)$ in the bottom right of Figure~\ref{fig:block_diagram} are examples of such combinations. We then execute each entry in $\mathbf{e}$ to get a list of states \state{}{}. This is followed by embedding each entry in states \state{}{} using $H_{\theta}$ to yield a tensor of state embeddings $\mathbfcal{H}$(\state{}{}) $\in \mathbb{R}^{L \times Z}$ (henceforth referred to as \statemb{}{} for simplicity). The white box towards the bottom of Figure~\ref{fig:block_diagram} shows an example of obtaining a single entry of a PE state embedding.

\textbf{Motivation:} To explain the motivation behind CA, let's look at Figure~\ref{fig:block_diagram}, which illustrates the process of synthesizing line 2 of $p_g$. Intuitively, at this step, we will want our aggregation mechanism to have more contribution from line 2 of $p_1$ and $p_3$ (i.e., \code{DROP c a}). A simple way of aggregating the PE solutions can be to take the sum or mean of the PE one-hot statement vectors (these form our ablation baselines as detailed in Section~\ref{sec:methods}). However, this strategy will fail for scenarios that require taking a non-trivial combination of the PE solution statements or cases where the global solution requires the generation of a new statement that is not found in the PE solutions. 

In this work, we propose another way of anticipating what line of $p_g$ comes next, that makes use of the execution information of the programs. The idea is to compare the state embedding obtained before executing line 2 of $p_g$ with the PE state embeddings corresponding to each step of execution of the PE solutions. Then, based on the learned relevance of these state embeddings, their corresponding next PE program statements can form valuable cues for synthesizing the next line. In other words, if a particular PE program state has high relevance with the global program state at a given step, then the following PE program line is likely to be useful in synthesizing the next line of $p_g$.  We measure this relevance by employing a cross-attention mechanism, with the query formed by the global program state embedding at step $t$, a key formed by the PE program state embedding at step $t$ and the corresponding value formed by the PE program statement at $t+1$. We take a set of such keys and values to form the key matrix $\mathbf{K}$ and the value matrix $\mathbf{V}$, respectively.

\textbf{Model:} For synthesizing line $t+1$ of $p_g$, the query $\mathbf{Q}$ is formed from the global state embedding at step $t$, denoted by \statemb{query}{t} $\in \mathbb{R}^{1 \times Z}$, where \etb{query}{g}{\{1, 2, \dots N\}}{t}. The keys $\mathbf{K} \in \mathbb{R}^{L \times Z}$ are formed from the state embeddings \statemb{keys}{} of the PE solutions. Let $P$ denote the list of $M$ discovered PE solutions, then the list of execution tuples \etb{keys}{m}{\{j\}}{}, where $p_m \in P, j \in \{1, 2, ..N\}$, $t \in \{0, 1, ..|p_m|-1\}$, making $L = M \times N \times \sum_{m=1}^{M} |p_m|$. The corresponding PE solution statements form the values $\mathbf{V} \in \mathbb{R}^{L \times Z}$ (more details on how values are obtained is given later). In addition, we have the relation scores $\mathbfcal{U} \in \mathbb{R}^{L \times 1}$ obtained by taking the PE solution score $u_m$ corresponding to $p_m$ that is part of each ET in $\mathbf{e_{keys}}$. Note that entries in $\mathbfcal{U}$ are dependent only on the program part in the ET, and independent of the subset of example indices and the time index.

We add position encodings (depending on the time step value of each ET) to $\mathbf{Q}$, $\mathbf{K}$ and $\mathbf{V}$. This is followed by multiheaded relative attention between our keys, values and query as described in Equation~\ref{eq:mha}. For each head, we perform a scaled dot-product attention~\cite{transformer}(Equation~\ref{eq:sda}) and a form of relative attention\footnote{Note that our formulation of relative attention differs from the formulation used in \citet{shaw-etal-2018-self, Hellendoorn2020Global}, where the relation scores are added either to the query or values.}, i.e. taking a mean of the relation scores and attention scores before normalizing with softmax and multiplying with values (Equation~\ref{eq:ra}). 

\begin{align}
    Att(\mathbf{Q}, \mathbf{K}) &= \frac{\mathbf{Q}\mathbf{K}^T}{\sqrt{d_k}} \label{eq:sda}\\
    RelAtt(\mathbf{Q}, \mathbf{K}, \mathbf{V}) &= \text{softmax} \Big(\frac{\mathbfcal{U}^T + Att(\mathbf{Q}, \mathbf{K})}{2}\Big) \mathbf{V}\label{eq:ra}\\
    MultiHead(\mathbf{Q}, \mathbf{K}, \mathbf{V}) &= \text{concat} (head_i, head_2,  \dots head_{\tau})W^{O} \label{eq:mha}\\ \nonumber
    \text{where} \quad  head_i &= RelAtt (\mathbf{Q}W_i^{Q}, \mathbf{K}W_i^{K},  \mathbf{V}W_i^{V}) 
\end{align}

In the equations above, $d_k$ is the dimension of the key, $W_i^{Q}, W_i^{K}, W_i^{V}$ are the query, key and value projection matrices, $\tau$ is the number of heads and $W^O$ is the linear projection that combines the heads. The output from Equation~\ref{eq:mha} is fed to a positionwise fully-connected feedforward network. We employ a residual connection~\cite{he2016deep} followed by layer normalization~\cite{ba2016layer} before and after the feedforward network. The resulting encoding is then linearly projected and softmax is applied to get the prediction of the statement for line $t+1$ of $p_g$. We see that our model resembles one layer of the transformer encoder block \cite{transformer}. Since the keys and query come from different sources, we refer to our model as a {\it cross} aggregator. Like standard transformers, we can stack multiple blocks of CA. However, since we are operating on a low timeout (5s), we opted for a simple network consisting of only one layer. Details of model parameters can be found in Appendix D.3.

\textbf{Obtaining $\mathbf{V}$:} For a key corresponding to an ET consisting of the PE solution $p_m$ and having step index $t$, the value is associated with the statement vector (one-hot vector of size = $n_s$) for step $t+1$ of $p_m$. Putting together the statement vectors for all execution tuples that are part of $\mathbf{e_{keys}}$, we get a tensor $\mathbf{p_{values}}$ of size $L \times n_s$.
Embedding each entry in this tensor using an embedding layer $F_{\gamma}$ gives us $\mathbf{V} = \mathbfcal{F}\mathbf{(p_{values})}$ of size $L \times Z$. This is then fed as input to the model described above. The output from the model is then linearly projected to give the logits for the statement predictions $\in \mathbb{R}^{n_s}$ for step $t+1$ of the global program $p_g$. In addition to the statement predictions, we can also obtain the operator predictions $\in \mathbb{R}^{n_o}$, starting from the operator vector (one-hot vector of size = $n_o$) and following a process similar to the statements, except that we use a different embedding and final projection layer. The right of Figure~\ref{fig:block_diagram} shows an example of how a query (top) is combined with keys, values and relation scores (bottom) for our model.

\subsection{Training}
\label{sec:train_peps}
The two main components of N-PEPS, the PE Searches module and the Cross-Aggregator, are trained separately. To create samples for training the PE model, we take one data point ($X=\{r_i\}_{i=1}^N$ and $p_g$) from the GPS approach and create $N$ data points out of it. Since we do not have supervision for the PE solutions, for every example $r_i$ in $X$, we use $p_g$ as a proxy for ground-truth PE solution. We believe that using $p_g$ as proxy supervision even though not being entirely correct, forces the PE search component to avoid overfitting to a single example and hence is more likely to produce PE solutions that generalize to examples outside the ones given as specification (see Appendix D.2 for more details).

For training the CA module, we generate data points that we call \emph{aggregator instances}. Each aggregator instance consists of $X$, a list $Y$ of tuples of PE solutions $p_i$ and corresponding PE solution scores $u_i$, and global program $p_g$. The $p_i$'s and $u_i$'s are generated via CAB from a trained PE model (more details on how they are generated in Appendix C.2). Given $X$ and $Y$ as input, the objective is to learn the parameters of the CA module such that the output is the line-wise statement and operator predictions corresponding to $p_g$. The net loss at step $t$ is the sum of two terms: (a) a cross entropy loss between the predicted statement $\hat{s}^t$ (obtained from CA) and the actual statement vector $s^t$ (obtained from $p_g^t$); (b) a cross entropy loss between the predicted operator $\hat{o}^t$ and the actual operator vector $o^t$. Like PCCoder, the operator loss is used as an auxiliary loss to improve training. Note that for each aggregator instance, since we have $X$ and $Y$ to begin with, we need to compute the keys and values only once. However, the computation of query has to be done at each step of the global program execution. While training, since $p_g$ is known, we can use teacher forcing and increase efficiency by batching, where an element in the batch corresponds of one step of execution of $p_g$. 

\subsection{Inference}
The process of inference in PEPS is the same as in PCCoder (see Section~\ref{sec:pccoder}), except that in addition to the contribution from GPS, we add another term that accounts for the contribution from CA. The contribution from GPS is obtained by using a \emph{GPS model} that is trained as in standard PCCoder. The net value of the predicted statement at step $t$ is then obtained by taking a weighted contribution from the statement predictions from the trained GPS model $\hat{s}_{1-\alpha}^t$ and the statement prediction from the trained CA module $\hat{s}_{\alpha}^t$. For predicting the drop vector $\hat{d}^t$, we take contributions only from GPS. When $\alpha=0$, our approach becomes equivalent to GPS. 

\begin{align} 
    \hat{s}^t &= \alpha * \hat{s}_{\alpha}^t + (1 - \alpha) * \hat{s}_{1-\alpha}^t \label{eq:peps_inf}\\ \nonumber
    \hat{d}^t &= \hat{d}_{1 - \alpha}^t
\end{align}
We perform CAB until we find a global solution or we exceed the specified timeout. The right part of Figure~\ref{fig:block_diagram} illustrates an example of the steps involved in synthesizing step 2 of $p_g$.

\section{Experiments and Results}\label{experiments}

Following prior work~\citep{balog2016deepcoder, zohar2018automatic}\footnote{\label{fn1}We used the implementation from PCCoder~\cite{zohar2018automatic}, at https://github.com/amitz25/PCCoder (MIT License) for data generation and obtaining results for PCCoder.}, we generate programs for training and testing, with each program consisting of five IO example pairs, i.e., $N = 5$. The data generation process ensures that there is no overlap between the training and test programs, with programs being functionally non-equivalent to programs of shorter or equivalent lengths (see Appendix~\ref{app:train_test_prog_gen} for more details). In the first set of experiments (henceforth referred to as \textbf{E1}), we generated 105036 training programs of length up to 4 (i.e., consisting of lengths 1, 2, 3, 4). For the second set of experiments (henceforth referred to as \textbf{E2}), we generated 189328 training programs of length up to 12. 10\% of the training data was used for validation. To ensure robustness and reproducibility of results, for each method, we carry out experiments over 30 different test splits, where each split contains 500 programs of a specific length. For E1, we generate test programs of length 4, and for E2 we generate programs of lengths 5, 8, 10, 12 and 14. We learn separate GPS models and PE models for E1 and E2. All GPS results were obtained using the original PCCoder implementation\footnoteref{fn1}. A notable difference in our experiments from PCCoder~\cite{zohar2018automatic} is that we consider a short timeout of 5s (in both E1 and E2, \emph{for all methods}), instead of 5000s and 10000s. This choice is representative of the timeout required for satisfactory user experience in program synthesis systems used in real-world interactive use-cases (such as FlashFill feature in Microsoft Excel~\cite{flashfill}). Given a particular timeout value, we record the \emph{Success Ratio}, which is the fraction of test samples that succeeded in finding a global solution.  

\subsection{Initial Experiment: Analysis of PE Solutions}
\label{sec:hypothesis_experiment}
The promise of PEPS is rooted in the assumption that it is much easier to find PE solutions than finding a global solution. In order to test this hypothesis and get an idea of the types of cues discovered by PEPS, we performed a set of analysis experiments using data from E1. Using the trained PE model to find PE solutions, we consider two variants. The first variant called $\mathbf{tot(k)}$ is similar to the strategy of finding PE solutions that we use in PEPS (Section~\ref{sec:peps}), where we search for PE solutions sequentially (in the order of examples in $X$) until the discovered PE solutions taken together satisfy $k$ examples in $X$ (where $k \leq 5$). This helps us understand how much the coverage ($=k$) from a list of PE solutions can be. In the second variant called $\mathbf{ind(k)}$, we record success by searching for PE solutions sequentially until we find an individual PE solution that satisfies $k$ out of $N$ examples in $X$. Here, the success ratio helps us assess how good a single PE solution is. In other words, can we rely solely on individual solutions or do we need to aggregate them? For the initial experiment, since no aggregation is done, we divide the timeout of 5s evenly amongst PE searches, i.e., each PE search gets $\frac{1}{5} \times 5 = 1s$ as the timeout value. For GPS, we use the trained GPS model with a timeout of 5s.

\begin{table}[hbt!]
    \vskip -0.1in
    \scriptsize
	\centering
	\caption{Success ratio of GPS, $\mathbf{ind}$ and $\mathbf{tot}$ for different values of $k$ for test programs of length 4.}
	\begin{tabular}{c|ccccc|ccccc}
	\toprule
	\textbf{GPS} & $\mathbf{ind(1)}$ & $\mathbf{ind(2)}$ & $\mathbf{ind(3)}$ & $\mathbf{ind(4)}$ & $\mathbf{ind(5)}$ & $\mathbf{tot(1)}$ & $\mathbf{tot(2)}$ & $\mathbf{tot(3)}$ & $\mathbf{tot(4)}$ & $\mathbf{tot(5)}$ \\
	\midrule
	77.0 & 99.2 & 95.4 & 85.4 & 70.4 & 43.2 & 99.2 & 97.6 & 97.0 & 94.8 & 82.4\\
	\bottomrule
	\end{tabular}
	\label{tab:hypothesis}
\end{table}

Table~\ref{tab:hypothesis} gives the results of these analysis experiments on one of test splits for programs of length 4. Note that in $\mathbf{tot}$, we are not aggregating the cues to find a global program. Hence, the value given under $\mathbf{tot(5)}$ is not directly comparable to GPS. We make a few observations. First, the success ratio increases with decreasing value of $k$. Therefore, as speculated, it is easier to find solutions that satisfy examples partially. Second, we see that even though the numbers for $\mathbf{ind}$ are encouraging, they are less than the corresponding values (for same $k$) for $\mathbf{tot}$. This suggests that aggregating PE solutions is better than dealing with them individually. Third, the success ratio of $\mathbf{tot(5)}$ is better than GPS. This suggests there is potential in thinking of an architecture that can learn to combine these solutions. Even for cases where we couldn't find PE solutions that satisfy all 5 examples, we can hope to make use of the rich partial cues (indicated by high success ratios) coming from $\mathbf{tot(k < 5)}$. 

\subsection{Methods}
\label{sec:methods}
In addition to the standard GPS baseline (PCCoder~\cite{zohar2018automatic}), we experimented with three ablation baselines that represent simple ways of aggregating the PE solutions without making use of the program state. Hence, they help us understand the role of PE cues alone. These baselines are: (i) \textbf{Sum-PEPS:} Replacing the contribution from CA module in Equation~\ref{eq:peps_inf} by a module that combines the PE solutions by taking the sum of all PE one-hot statement vectors; (ii) \textbf{Mean-PEPS:} Same as (i) except that sum is replaced by mean; (iii) \textbf{Mean-PEPS$+\mathbfcal{U}$:} Same as (ii) except that the one-hot PE statement vectors are multiplied by their corresponding solution scores before taking the mean. To understand the benefit of aggregating with our proposed CA architecture on top of the value brought by the PE cues, we experimented with the following variations: (i) \textbf{N-PEPS:} Our neural model of PEPS described in Section~\ref{sec: ca} with $\mathbfcal{U}$ being a zero tensor; (ii) \textbf{N-PEPS$+\mathbfcal{U}$:} Same as (i) but with $\mathbfcal{U}$ included. Complete details of hyperparameters for all methods can be found in Appendix~\ref{app:hyperparams}. 

\subsection{Results}
\label{sec:results}
For each test data point, we record either a success or failure (based on whether within 5s, we find a global solution or not) and the actual time taken to find a global solution. As described in Section~\ref{sec:peps}, for all PEPS methods, we start by allocating a PEPS timeout value that is less than 1s ($=\frac{1}{N} \times$ total timeout). We sum the actual time ($\leq$ PEPS timeout) taken for finding individual PE solutions. The residual times (if any) left out is added and used for aggregation and global inference. Note that PEPS timeout and $\alpha$ are treated as hyperparameters that are chosen using the validation set. 

To provide fair comparison across all methods, each test split is run using a single core and single CPU thread with a timeout of 5s. To account for variability across machines, we chose to run a test split on a machine chosen randomly from a collection of 7 machines of similar configuration (Google Cloud instances with 120GB RAM each)\footnote{We additionally verify that different runs on the same machine produce similar results (Appendix~\ref{app:ins_run_variation})}. We report standard error across the 30 test runs. 

\begin{figure}[t]
\begin{subfigure}{.33\textwidth}
\scriptsize
\begin{tabular}{cc}
	    \toprule
    	Model & Success Ratio\\
        \midrule
        PCCoder~\cite{zohar2018automatic} & 77.75 $\pm$ 0.38\\ 
        \midrule
        Sum-PEPS & 82.71 $\pm$ 0.32\\ 
        Mean-PEPS &  82.68 $\pm$ 0.33\\
		Mean-PEPS$+\mathcal{U}$ & 82.70 $\pm$ 0.32\\ 
		N-PEPS & 86.22 $\pm$ 0.25\\ 
		N-PEPS$+\mathcal{U}$ & \textbf{87.07 $\pm$ 0.28}\\ 
		\bottomrule
	\end{tabular}
\end{subfigure}
\hfill
\begin{subfigure}{.24\textwidth}
\includegraphics[width=1.3\linewidth, center]{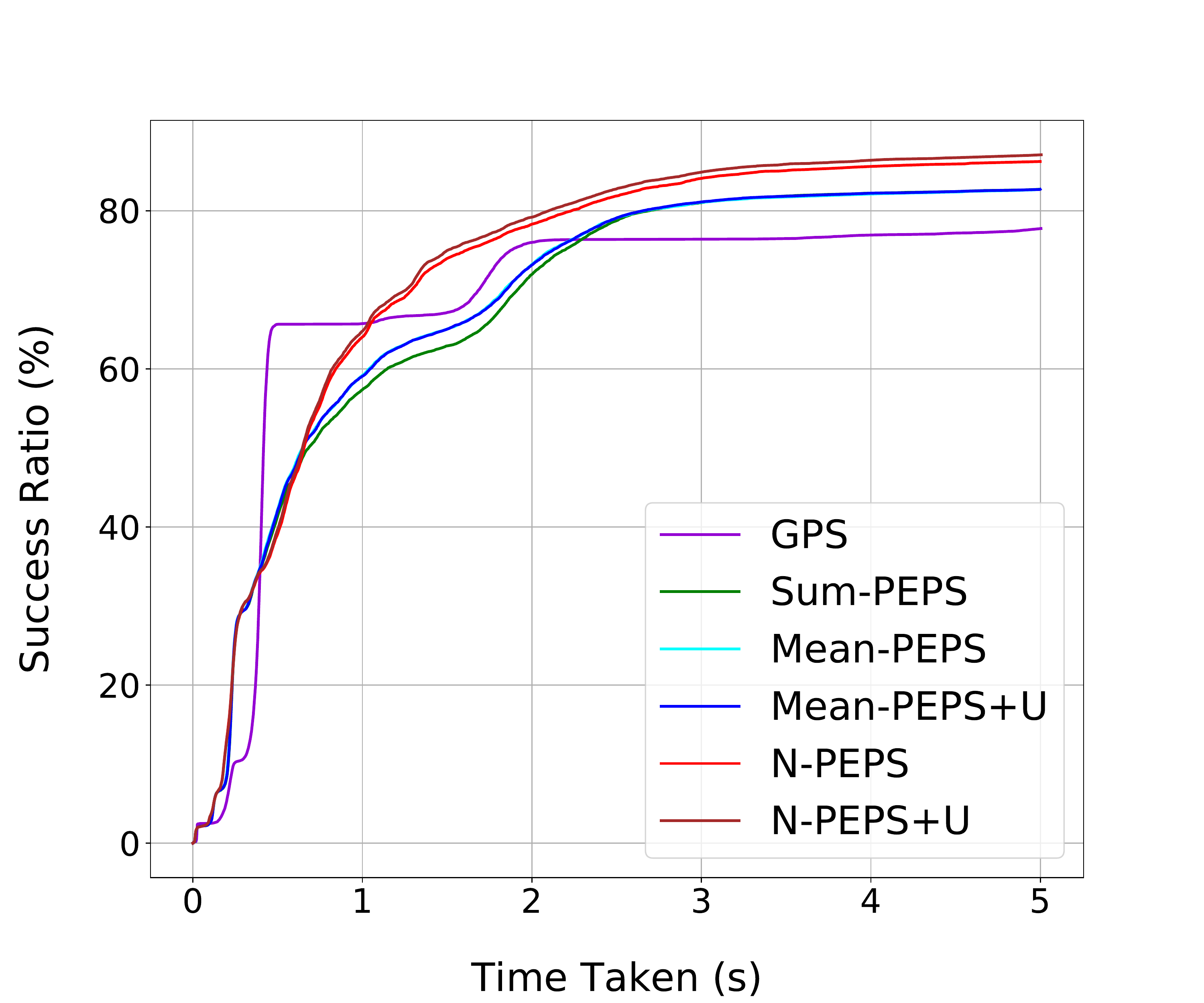}
\end{subfigure}
\hfill
\begin{subfigure}{.41\textwidth}
\includegraphics[width=0.9\linewidth, right]{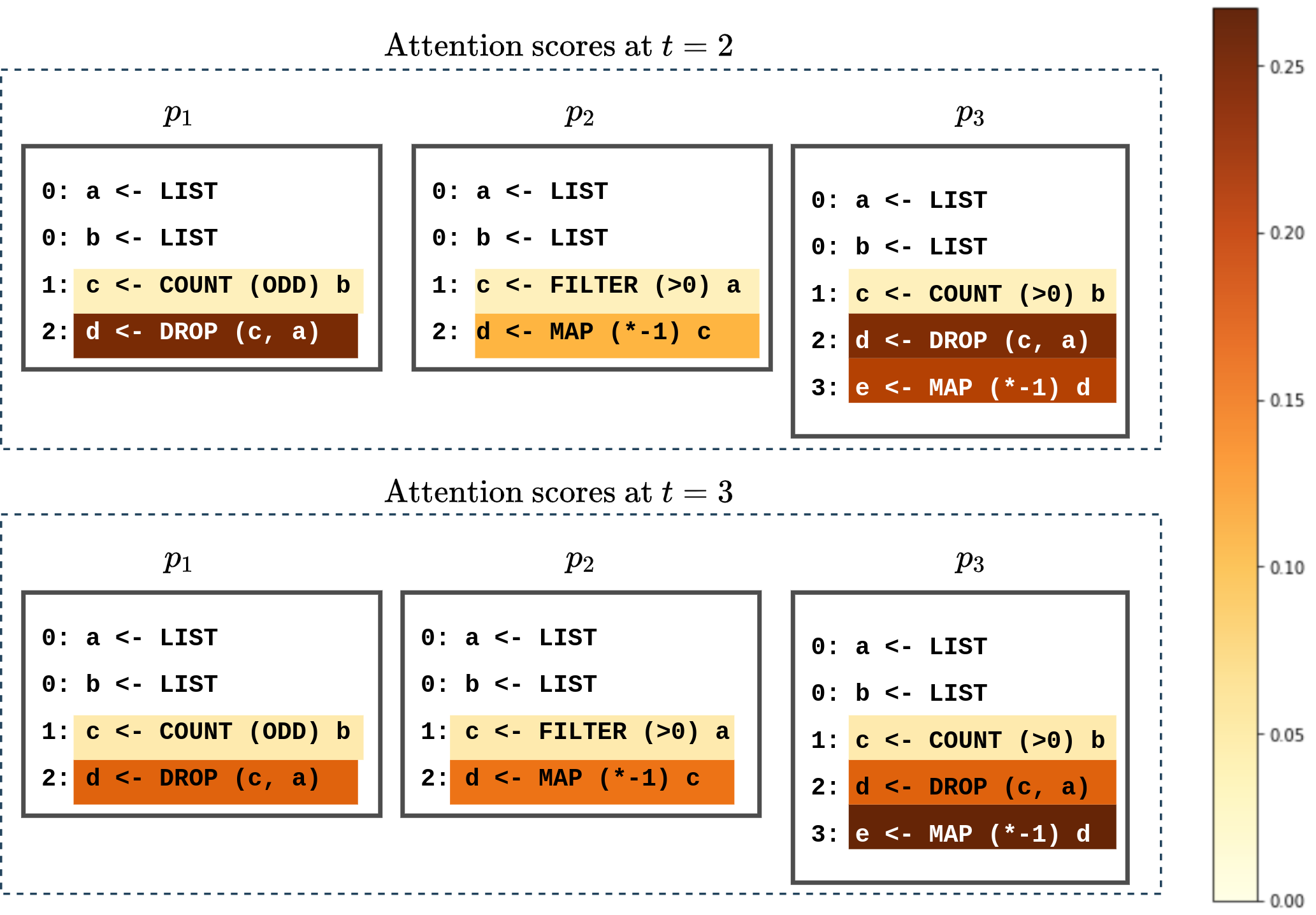}
\end{subfigure}
\caption{\textbf{Results for E1}: \textbf{\textit{(Left)}} Success Ratio with standard error for all models (top row = GPS); \textbf{\textit{(Center)}} Success Ratio vs. time taken; \textbf{\textit{(Right)}} Visualization of attention scores for N-PEPS$+\mathcal{U}$}
\label{fig:len_4}
\vskip -0.1in
\end{figure}

\begin{figure}[t]
\begin{subfigure}{.3\textwidth}
    \centering
    \scriptsize
	\begin{tabular}{cccccc}
	\toprule
	Model & Length = 5 & Length = 8 & Length = 10 & Length = 12 & Length=14\\
	\midrule
        PCCoder~\cite{zohar2018automatic} & 70.91 $\pm$ 0.35 & 44.17 $\pm$ 0.45 & 28.18 $\pm$ 0.33 & 19.69 $\pm$ 0.34 & 14.71 $\pm$ 0.23\\
    \midrule    
        Sum-PEPS & 76.45 $\pm$ 0.33 & 43.4 $\pm$ 0.56 & 28.96 $\pm$ 0.27 & 20.94 $\pm$ 0.32 & 15.67 $\pm$ 0.32\\

        Mean-PEPS & 75.79 $\pm$ 0.31 & 44.42 $\pm$ 0.51 & 29.55 $\pm$ 0.29 & 21.45 $\pm$ 0.27 & 16.35 $\pm$ 0.27\\
 
		Mean-PEPS$+\mathcal{U}$ & 75.99 $\pm$ 0.32 & 44.49 $\pm$ 0.52 & 29.75 $\pm$ 0.25 & 21.74 $\pm$ 0.30 & 16.45 $\pm$ 0.33\\ 
		
		N-PEPS & 79.18 $\pm$ 0.31 & \textbf{47.23 $\pm$ 0.49} & \textbf{32.3 $\pm$ 0.34} & \textbf{23.34 $\pm$ 0.28} & \textbf{17.35 $\pm$ 0.31}\\ 

		N-PEPS$+\mathcal{U}$ & \textbf{79.19 $\pm$ 0.30} & 46.31 $\pm$ 0.61 & 31.84 $\pm$ 0.36 & 22.71 $\pm$ 0.28 & 16.68 $\pm$ 0.21\\ 
		
	\bottomrule
	\end{tabular}
\end{subfigure}
\hfill
\begin{subfigure}{.27\textwidth}
 \includegraphics[width=0.9\linewidth, right]{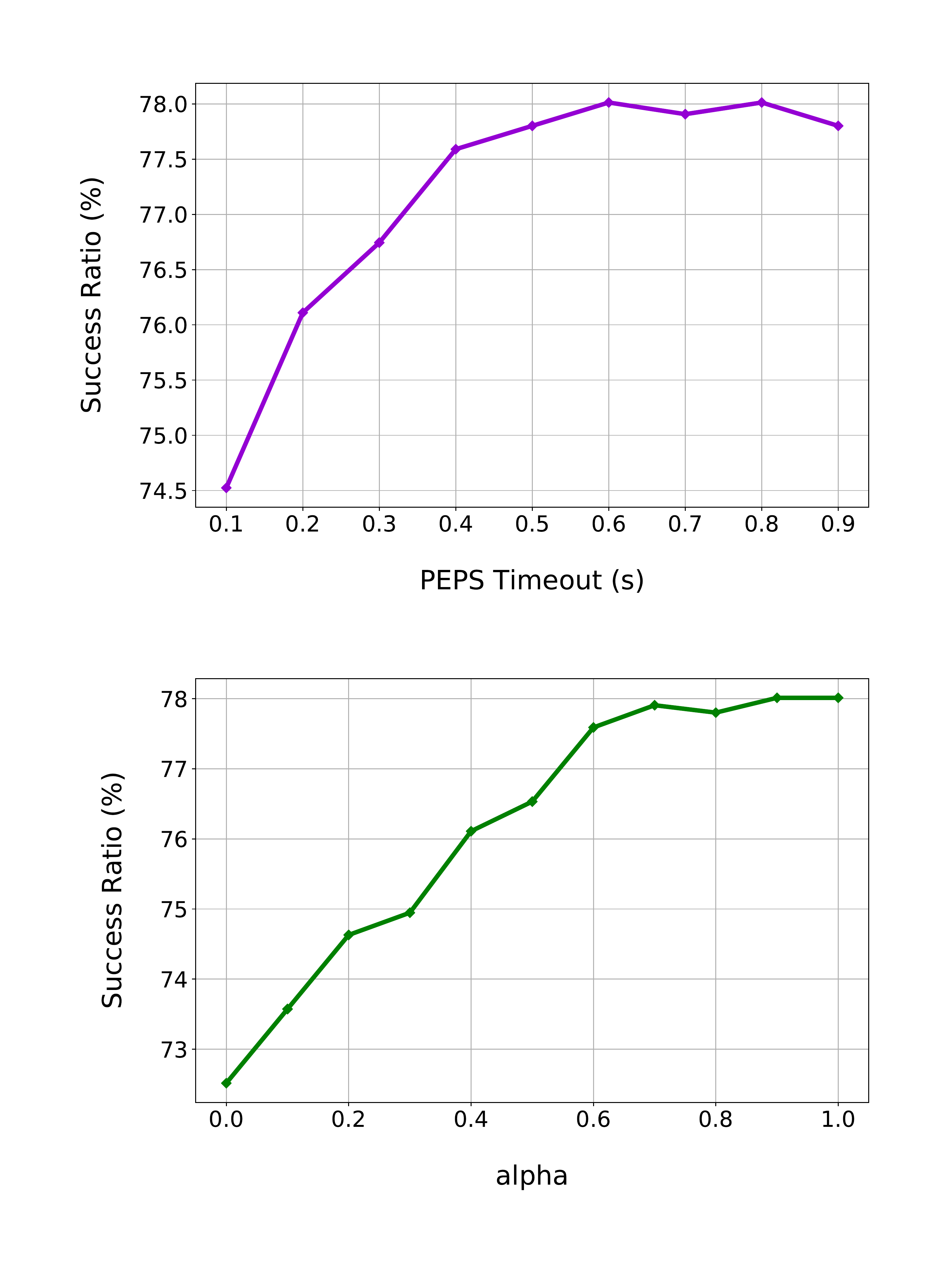}
\end{subfigure}
\vskip -0.1in
\caption{\textbf{Results for E2}: \textbf{\textit{(Left)}} Success Ratio with standard error for all models (top row = GPS); \textbf{\textit{(Right)}} Variation of success ratio with PEPS Timeout (top) and $\alpha$ (below) for N-PEPS}
\label{fig:len_12}
\vskip -0.1in
\end{figure}

\textbf{Main result:} The left and center parts of Figure~\ref{fig:len_4} show the success ratio and success ratio vs. time taken (average of the actual time taken) plots, respectively, for test programs of length 4 when trained on programs up to length 4 (E1). The left part of Figure~\ref{fig:len_12} shows the success ratio for test programs of lengths 5, 8, 10, 12 and 14, when trained on programs up to length 12 (E2). In both these settings, we observe that performance of the ablation baselines is better than GPS, illustrating the promise in the quality of PE solutions. When we use our CA module to aggregate these cues instead, we see that the performance improves even further. We used the default value of $\nu=11$ used in PCCoder, which means that for programs of length > 8, certain variables will be dropped from the program state. Also, note that the results for test length 14 represent a case of \emph{length generalization}. We show that in both these scenarios, our proposed method is quite advantageous\footnote{See Appendix~\ref{app:sample_cases} for success cases of N-PEPS \& Appendix~\ref{app:func_perf} and Appendix~\ref{app:perf_soln} for empirical analysis of synthesized programs.}.
In addition, we compare the performance of N-PEPS with GPS for the cases of \emph{intent generalization}, i.e., generalization of the synthesized program to examples other than those given as part of $X$ (Appendix~\ref{app:intent_gen}) and when given a longer timeout of 1000s (Appendix~\ref{app:longer_timeout}). In both these settings, N-PEPS shows superior performance, highlighting its generality.

\textbf{Attention visualization:} The right part of Figure~\ref{fig:len_4} shows a visualization of attention scores at $t=2$ and $t=3$ obtained from our best model under E1 for the example shown in Figure~\ref{fig:block_diagram}. This example represents a case from the test set where N-PEPS succeeds in finding a global solution, whereas other methods fail. As can be seen, the actual statement of $p_g^2$ is \code{DROP c a} and our model indeed gives relatively higher attention scores to $p_1^2$ and $p_3^2$, both of which correspond to the same statement. Similarly at $t=3$, our model gives more attention to $p_3^3 = \code{MAP (*-1) d} = p_g^3$.

\textbf{Variation with PEPS timeout and} $\boldsymbol\alpha$: There is a non-trivial tradeoff in the division of the total timeout into the time given to find PE solutions and the time given to the CA module. A higher PEPS timeout results in better chances of discovering rich PE cues. This means that there may be less or almost no time needed for aggregation. On the other hand, if we start with a low PEPS timeout, the cues from PE solutions may not be as good, but we have more time to perform aggregation. Also, there is a question of how much contribution should be taken from CA and how much from GPS, which is determined by the value of $\alpha$. The right part of Figure~\ref{fig:len_12} analyzes this tradeoff. In the top, we show the variation of success ratio with PEPS timeout and in the bottom, we have the variation with $\alpha$, for the validation set under E2. We see that the performance improves with increase in PEPS timeout with slight decrease towards the end. Also, we see that generally higher values of $\alpha$ are better indicating that the contribution from CA is more important than the contribution from GPS.

\begin{wraptable}{r}{3.5cm}
\scriptsize
\caption{Performance of variants of $\mathbf{K}$ for E1}
\vskip -0.1in
\label{tab:key_var}
\begin{tabular}{cc}\\
\toprule  
Variant & Success Ratio \\
\midrule
N-PEPS-PG & 85.19 $\pm$ 0.26\\ 
N-PEPS-PG+$\mathcal{U}$ & 85.94 $\pm$ 0.26\\ 
N-PEPS-PP & 85.97 $\pm$ 0.26\\ 
N-PEPS-PP+$\mathcal{U}$ & 86.21 $\pm$ 0.27\\
N-PEPS & 86.22 $\pm$ 0.25\\ 
N-PEPS$+\mathcal{U}$ & \textbf{87.07 $\pm$ 0.28}\\ 
\bottomrule
\end{tabular}
\vskip -0.1in
\end{wraptable} 

\textbf{Variants of $\mathbf{K}$:} In addition to obtaining \statemb{keys}{} in the way described in Section~\ref{sec: ca}, we tried two other ways of composing the keys by varying the set $\mathcal{S}$ in the execution tuple against which the PE solutions are executed. In the first variant, PE solution $p_m$ from the list $P$ of $M$ discovered PE solutions is executed against the global set $X$, i.e., \etb{keys}{m}{\{1, 2, ..N\}}{} where $p_m \in P$ and $t \in \{0, 1, ..|p_m|-1\}$. We denote this variant as \textbf{N-PEPS-PG} (PG = PE-global ET). In the second variant, $p_m$ is executed against the set $S_m$ consisting only of examples indices that $p_m$ satisfies, i.e., \etb{keys}{m}{\{j\}}{} where $j \in  S_m$. We call this variant as \textbf{N-PEPS-PP} (PP = PE-PE ET). Table~\ref{tab:key_var} shows the test results of these variants for E1 with and without $\mathcal{U}$. We see that all the variants perform better than GPS and the three ablation baselines (N-PEPS variant used in Section~\ref{sec: ca} was chosen using the validation set). We see similar trend for E2 (see Appendix~\ref{app:key_var_12}).

\textbf{New operator discovery by N-PEPS:} We were interested in determining that while synthesizing the global solution how often does N-PEPS rely on copying statements from the PE solutions and how often does it generate new operators. We studied this trend for the cases when $\alpha < 1.0$, i.e., contributions are taken from both CA and GPS as well as when $\alpha=1.0$, i.e., contributions are taken only from CA (Appendix~\ref{app:overlap_pe_global}). This question is important as it helps us understand the generalization capabilities of CA outside the statements in the PE solutions. We found that CA alone (with $\alpha=1.0$) is capable of generating new operators. In addition, we found that the new operators are present as part of the nearest neighbours of the PE statements, thereby pointing to an increased likelihood of these being ranked higher in the beam search and hence being present in the global solution (see Appendix~\ref{app:sample_new_gen} and \ref{app:nn} for details).

\section{Related Work}\label{related_work}

There have been numerous efforts on using deep learning for program synthesis~\cite{balog2016deepcoder, bunel2018leveraging, robustfill, kalyan2018neuralguided, npmi, sygus, sketches19, Odena2020Learning, parisotto2016neurosymbolic}. However, there is less work that uses the execution of partial programs to assist in synthesis. PCCoder~\cite{zohar2018automatic} is one such work, which we describe in Section~\ref{sec:pccoder}. BUSTLE~\cite{odena2021bustle} reweighs the sub-expressions in bottom-up program synthesis using the intermediate values obtained by execution of sub-expressions along with property signatures. REPL~\cite{REPL19} executes partial programs using learned policy and value networks. \citet{chen2018execution} uses a neural encoder-decoder architecture to generate program tokens conditioned on intermediate states obtained from execution of partial programs. They work with the Karel DSL~\cite{karel, bunel2018leveraging} that contains loops and conditionals, an attribute missing from the DSL which we work with. Therefore, extending N-PEPS for Karel is an interesting future work. Note that all the approaches mentioned above are examples of purely GPS approaches.

Few works use solutions that satisfy examples partially, to aid in program synthesis. Initial motivation for our work comes from FrAngel~\cite{shi2019frangel}, which is a component-wise synthesis system that relies on mining fragments of Java code that satisfy examples partially, given target program function signatures and a list of Java libraries. The mining of fragments as well as combination is done using a set of heuristics and predefined rules with no deep learning involved. Assuming that the user provides IO examples in order of increasing difficulty, \citet{perelman2014test} iteratively refines a program, with the current program satisfying the sequence of IO examples encountered till now. STUN~\cite{alur2015synthesis} extends the CEGIS~\cite{DBLP:conf/asplos/Solar-LezamaTBSS06} approach by providing domain-specific explicit unification operators for combining partial solutions while \citet{Alur2017ScalingEP} uses decision trees for the same. Recently, BESTER~\cite{peleg2020perfect} and later PROBE~\cite{barke2020just} perform bottom-up enumeration of programs in a loop by enumerating all programs that satisfy IO examples partially. This is followed by heuristics-based selection of promising programs. However, as opposed to N-PEPS that automatically learns to aggregate partial solutions producing the global program in one shot, PROBE relies on using these programs to iteratively update the weights of useful productions in their probabilistic grammar using a fixed update rule. This update can be viewed similar to our ablation baselines that do not use the neural network based learned aggregation. The guided-search component of PROBE provides an interesting alternative to finding PE solutions. One way of incorporating this in our top-down setting might be to start with the CAB search as in GPS and then select promising solutions based on evaluating examples on prefixes of programs obtained during the beam search. It may be useful to then aggregate the selected solutions using a neural architecture similar to ours.

\section{Conclusions and Future Directions}
\label{conclusion}

In this work, we propose N-PEPS, where the idea is to break the problem of finding a program that solves all examples into two stages: (a) finding programs that solve a single example (PE solutions) (b) aggregating the PE solutions such that it leads to a program that solves all examples. For aggregation, we propose a neural-network based multi-head attention architecture (CA module) that utilizes the state of program execution to learn to combine the PE cues. We note that program synthesis systems in general should be deployed with caution for use in real-world applications. Blind trust on these systems can create chances for potential negative impact, as there might be cases where the generated program contains bugs, especially for unseen examples outside the specification. In the future, we want to work with programs that contain loops and conditionals~\cite{karel, chen2018execution, bunel2018leveraging}. Another interesting research direction would be to explore the interaction of N-PEPS with out-of-distribution generalization settings like compositional generalization~\cite{scan}.

\begin{ack}
Hugo Larochelle would like to acknowledge the support of Canada Research Chairs and CIFAR for research funding. The authors would like to thank Google Cloud for providing compute resources required for this project. We would like to thank the anonymous reviewers for their valuable comments and thorough engagement during the rebuttal phase that helped us improve our paper. We would also like to extend our vote of thanks to Daniel Johnson, Petros Maniatis, Jacob Austin, Koushik Sen, David Bieber, Sandeep Subramanian, Varsha Embar, Nicolas Gontier and Andreea Deac for feedback and comments on the draft that helped us improve writing. We would also like to acknowledge the quick and helpful response from Amit Zohar and Lior Wolf on queries related to the PCCoder implementation.
\end{ack}

\bibliographystyle{plainnat}{
\bibliography{neurips_2021}

\begin{thebibliography}{32}
\providecommand{\natexlab}[1]{#1}
\providecommand{\url}[1]{\texttt{#1}}
\expandafter\ifx\csname urlstyle\endcsname\relax
  \providecommand{\doi}[1]{doi: #1}\else
  \providecommand{\doi}{doi: \begingroup \urlstyle{rm}\Url}\fi

\bibitem[Alur et~al.(2015)Alur, {\v{C}}ern{\`y}, and
  Radhakrishna]{alur2015synthesis}
Rajeev Alur, Pavol {\v{C}}ern{\`y}, and Arjun Radhakrishna.
\newblock Synthesis through unification.
\newblock In \emph{International Conference on Computer Aided Verification},
  pages 163--179. Springer, 2015.

\bibitem[Alur et~al.(2017)Alur, Radhakrishna, and Udupa]{Alur2017ScalingEP}
Rajeev Alur, Arjun Radhakrishna, and Abhishek Udupa.
\newblock Scaling enumerative program synthesis via divide and conquer.
\newblock In \emph{TACAS}, 2017.

\bibitem[Ba et~al.(2016)Ba, Kiros, and Hinton]{ba2016layer}
Jimmy~Lei Ba, Jamie~Ryan Kiros, and Geoffrey~E. Hinton.
\newblock Layer normalization, 2016.

\bibitem[Balog et~al.(2016)Balog, Gaunt, Brockschmidt, Nowozin, and
  Tarlow]{balog2016deepcoder}
Matej Balog, Alexander~L Gaunt, Marc Brockschmidt, Sebastian Nowozin, and
  Daniel Tarlow.
\newblock Deepcoder: Learning to write programs.
\newblock In \emph{International Conference on Learning Representations}, 2016.

\bibitem[Barke et~al.(2020)Barke, Peleg, and Polikarpova]{barke2020just}
Shraddha Barke, Hila Peleg, and Nadia Polikarpova.
\newblock Just-in-time learning for bottom-up enumerative synthesis.
\newblock \emph{Proceedings of the ACM on Programming Languages}, 4\penalty0
  (OOPSLA):\penalty0 1--29, 2020.

\bibitem[Bunel et~al.(2018)Bunel, Hausknecht, Devlin, Singh, and
  Kohli]{bunel2018leveraging}
Rudy Bunel, Matthew Hausknecht, Jacob Devlin, Rishabh Singh, and Pushmeet
  Kohli.
\newblock Leveraging grammar and reinforcement learning for neural program
  synthesis.
\newblock In \emph{International Conference on Learning Representations}, 2018.

\bibitem[Chen et~al.(2018)Chen, Liu, and Song]{chen2018execution}
Xinyun Chen, Chang Liu, and Dawn Song.
\newblock Execution-guided neural program synthesis.
\newblock In \emph{International Conference on Learning Representations}, 2018.

\bibitem[Devlin et~al.(2017{\natexlab{a}})Devlin, Bunel, Singh, Hausknecht, and
  Kohli]{npmi}
Jacob Devlin, Rudy~R Bunel, Rishabh Singh, Matthew Hausknecht, and Pushmeet
  Kohli.
\newblock {Neural Program Meta-Induction}.
\newblock In I~Guyon, U~V Luxburg, S~Bengio, H~Wallach, R~Fergus,
  S~Vishwanathan, and R~Garnett, editors, \emph{Advances in Neural Information
  Processing Systems}, volume~30. Curran Associates, Inc., 2017{\natexlab{a}}.

\bibitem[Devlin et~al.(2017{\natexlab{b}})Devlin, Uesato, Bhupatiraju, Singh,
  Mohamed, and Kohli]{robustfill}
Jacob Devlin, Jonathan Uesato, Surya Bhupatiraju, Rishabh Singh, Abdel-rahman
  Mohamed, and Pushmeet Kohli.
\newblock Robustfill: Neural program learning under noisy i/o.
\newblock In \emph{Proceedings of the 34th International Conference on Machine
  Learning - Volume 70}, ICML'17, page 990–998. JMLR.org, 2017{\natexlab{b}}.

\bibitem[Ellis et~al.(2019)Ellis, Nye, Pu, Sosa, Tenenbaum, and
  Solar-Lezama]{REPL19}
Kevin Ellis, Maxwell Nye, Yewen Pu, Felix Sosa, Josh Tenenbaum, and Armando
  Solar-Lezama.
\newblock Write, execute, assess: Program synthesis with a repl.
\newblock In H.~Wallach, H.~Larochelle, A.~Beygelzimer, F.~d\textquotesingle
  Alch\'{e}-Buc, E.~Fox, and R.~Garnett, editors, \emph{Advances in Neural
  Information Processing Systems}, volume~32. Curran Associates, Inc., 2019.

\bibitem[Gulwani(2011)]{flashfill}
Sumit Gulwani.
\newblock Automating string processing in spreadsheets using input-output
  examples.
\newblock In \emph{Proceedings of the 38th Annual ACM SIGPLAN-SIGACT Symposium
  on Principles of Programming Languages}, POPL '11, page 317–330, New York,
  NY, USA, 2011. Association for Computing Machinery.
\newblock ISBN 9781450304900.
\newblock \doi{10.1145/1926385.1926423}.

\bibitem[Gupta et~al.(2020)Gupta, Christensen, Chen, and Song]{sed}
Kavi Gupta, Peter~Ebert Christensen, Xinyun Chen, and Dawn Song.
\newblock Synthesize, execute and debug: Learning to repair for neural program
  synthesis.
\newblock In H.~Larochelle, M.~Ranzato, R.~Hadsell, M.~F. Balcan, and H.~Lin,
  editors, \emph{Advances in Neural Information Processing Systems}, volume~33,
  pages 17685--17695. Curran Associates, Inc., 2020.

\bibitem[He et~al.(2016)He, Zhang, Ren, and Sun]{he2016deep}
Kaiming He, Xiangyu Zhang, Shaoqing Ren, and Jian Sun.
\newblock Deep residual learning for image recognition.
\newblock In \emph{Proceedings of the IEEE conference on computer vision and
  pattern recognition}, pages 770--778, 2016.

\bibitem[Hellendoorn et~al.(2020)Hellendoorn, Sutton, Singh, Maniatis, and
  Bieber]{Hellendoorn2020Global}
Vincent~J. Hellendoorn, Charles Sutton, Rishabh Singh, Petros Maniatis, and
  David Bieber.
\newblock Global relational models of source code.
\newblock In \emph{International Conference on Learning Representations}, 2020.

\bibitem[Kalyan et~al.(2018)Kalyan, Mohta, Polozov, Batra, Jain, and
  Gulwani]{kalyan2018neuralguided}
Ashwin Kalyan, Abhishek Mohta, Oleksandr Polozov, Dhruv Batra, Prateek Jain,
  and Sumit Gulwani.
\newblock Neural-guided deductive search for real-time program synthesis from
  examples.
\newblock In \emph{International Conference on Learning Representations}, 2018.

\bibitem[Kingma and Ba(2015)]{DBLP:journals/corr/KingmaB14}
Diederik~P. Kingma and Jimmy Ba.
\newblock Adam: {A} method for stochastic optimization.
\newblock In Yoshua Bengio and Yann LeCun, editors, \emph{3rd International
  Conference on Learning Representations, {ICLR} 2015, San Diego, CA, USA, May
  7-9, 2015, Conference Track Proceedings}, 2015.

\bibitem[Lake and Baroni(2018)]{scan}
Brenden Lake and Marco Baroni.
\newblock Generalization without systematicity: On the compositional skills of
  sequence-to-sequence recurrent networks.
\newblock In Jennifer Dy and Andreas Krause, editors, \emph{Proceedings of the
  35th International Conference on Machine Learning}, volume~80 of
  \emph{Proceedings of Machine Learning Research}, pages 2873--2882. PMLR,
  10--15 Jul 2018.

\bibitem[Lee et~al.(2018)Lee, Heo, Alur, and Naik]{sygus}
Woosuk Lee, Kihong Heo, Rajeev Alur, and Mayur Naik.
\newblock Accelerating search-based program synthesis using learned
  probabilistic models.
\newblock In \emph{Proceedings of the 39th ACM SIGPLAN Conference on
  Programming Language Design and Implementation}, PLDI 2018, page 436–449,
  New York, NY, USA, 2018. Association for Computing Machinery.
\newblock ISBN 9781450356985.
\newblock \doi{10.1145/3192366.3192410}.

\bibitem[Nye et~al.(2019)Nye, Hewitt, Tenenbaum, and Solar-Lezama]{sketches19}
Maxwell Nye, Luke Hewitt, Joshua Tenenbaum, and Armando Solar-Lezama.
\newblock Learning to infer program sketches.
\newblock In Kamalika Chaudhuri and Ruslan Salakhutdinov, editors,
  \emph{Proceedings of the 36th International Conference on Machine Learning},
  volume~97 of \emph{Proceedings of Machine Learning Research}, pages
  4861--4870. PMLR, 09--15 Jun 2019.

\bibitem[Odena and Sutton(2020)]{Odena2020Learning}
Augustus Odena and Charles Sutton.
\newblock Learning to represent programs with property signatures.
\newblock In \emph{International Conference on Learning Representations}, 2020.

\bibitem[Odena et~al.(2021)Odena, Shi, Bieber, Singh, Sutton, and
  Dai]{odena2021bustle}
Augustus Odena, Kensen Shi, David Bieber, Rishabh Singh, Charles Sutton, and
  Hanjun Dai.
\newblock {\{}BUSTLE{\}}: Bottom-up program synthesis through learning-guided
  exploration.
\newblock In \emph{International Conference on Learning Representations}, 2021.

\bibitem[Parisotto et~al.(2016)Parisotto, rahman Mohamed, Singh, Li, Zhou, and
  Kohli]{parisotto2016neurosymbolic}
Emilio Parisotto, Abdel rahman Mohamed, Rishabh Singh, Lihong Li, Dengyong
  Zhou, and Pushmeet Kohli.
\newblock Neuro-symbolic program synthesis, 2016.

\bibitem[Paszke et~al.(2019)Paszke, Gross, Massa, Lerer, Bradbury, Chanan,
  Killeen, Lin, Gimelshein, Antiga, Desmaison, Kopf, Yang, DeVito, Raison,
  Tejani, Chilamkurthy, Steiner, Fang, Bai, and Chintala]{NEURIPS2019_9015}
Adam Paszke, Sam Gross, Francisco Massa, Adam Lerer, James Bradbury, Gregory
  Chanan, Trevor Killeen, Zeming Lin, Natalia Gimelshein, Luca Antiga, Alban
  Desmaison, Andreas Kopf, Edward Yang, Zachary DeVito, Martin Raison, Alykhan
  Tejani, Sasank Chilamkurthy, Benoit Steiner, Lu~Fang, Junjie Bai, and Soumith
  Chintala.
\newblock Pytorch: An imperative style, high-performance deep learning library.
\newblock In H.~Wallach, H.~Larochelle, A.~Beygelzimer, F.~d\textquotesingle
  Alch\'{e}-Buc, E.~Fox, and R.~Garnett, editors, \emph{Advances in Neural
  Information Processing Systems 32}, pages 8024--8035. Curran Associates,
  Inc., 2019.

\bibitem[Pattis(1994)]{karel}
Richard~E. Pattis.
\newblock \emph{Karel the Robot: A Gentle Introduction to the Art of
  Programming}.
\newblock John Wiley \& Sons, Inc., USA, 2nd edition, 1994.
\newblock ISBN 0471107026.

\bibitem[Peleg and Polikarpova(2020)]{peleg2020perfect}
Hila Peleg and Nadia Polikarpova.
\newblock Perfect is the enemy of good: Best-effort program synthesis.
\newblock In \emph{34th European Conference on Object-Oriented Programming
  (ECOOP 2020)}. Schloss Dagstuhl-Leibniz-Zentrum f{\"u}r Informatik, 2020.

\bibitem[Perelman et~al.(2014)Perelman, Gulwani, Grossman, and
  Provost]{perelman2014test}
Daniel Perelman, Sumit Gulwani, Dan Grossman, and Peter Provost.
\newblock Test-driven synthesis.
\newblock \emph{ACM Sigplan Notices}, 49\penalty0 (6):\penalty0 408--418, 2014.

\bibitem[Shaw et~al.(2018)Shaw, Uszkoreit, and Vaswani]{shaw-etal-2018-self}
Peter Shaw, Jakob Uszkoreit, and Ashish Vaswani.
\newblock Self-attention with relative position representations.
\newblock In \emph{Proceedings of the 2018 Conference of the North {A}merican
  Chapter of the Association for Computational Linguistics: Human Language
  Technologies, Volume 2 (Short Papers)}, pages 464--468, New Orleans,
  Louisiana, June 2018. Association for Computational Linguistics.
\newblock \doi{10.18653/v1/N18-2074}.

\bibitem[Shi et~al.(2019)Shi, Steinhardt, and Liang]{shi2019frangel}
Kensen Shi, Jacob Steinhardt, and Percy Liang.
\newblock Frangel: component-based synthesis with control structures.
\newblock \emph{Proceedings of the ACM on Programming Languages}, 3\penalty0
  (POPL):\penalty0 1--29, 2019.

\bibitem[Solar{-}Lezama et~al.(2006)Solar{-}Lezama, Tancau, Bod{\'{\i}}k,
  Seshia, and Saraswat]{DBLP:conf/asplos/Solar-LezamaTBSS06}
Armando Solar{-}Lezama, Liviu Tancau, Rastislav Bod{\'{\i}}k, Sanjit~A. Seshia,
  and Vijay~A. Saraswat.
\newblock Combinatorial sketching for finite programs.
\newblock In John~Paul Shen and Margaret Martonosi, editors, \emph{Proceedings
  of the 12th International Conference on Architectural Support for Programming
  Languages and Operating Systems, {ASPLOS} 2006, San Jose, CA, USA, October
  21-25, 2006}, pages 404--415. {ACM}, 2006.
\newblock \doi{10.1145/1168857.1168907}.

\bibitem[Vaswani et~al.(2017)Vaswani, Shazeer, Parmar, Uszkoreit, Jones, Gomez,
  Kaiser, and Polosukhin]{transformer}
Ashish Vaswani, Noam Shazeer, Niki Parmar, Jakob Uszkoreit, Llion Jones,
  Aidan~N. Gomez, undefinedukasz Kaiser, and Illia Polosukhin.
\newblock Attention is all you need.
\newblock In \emph{Proceedings of the 31st International Conference on Neural
  Information Processing Systems}, NIPS'17, page 6000–6010, Red Hook, NY,
  USA, 2017. Curran Associates Inc.
\newblock ISBN 9781510860964.

\bibitem[Zhang(1998)]{cab}
Weixiong Zhang.
\newblock Complete anytime beam search.
\newblock In \emph{Proceedings of the Fifteenth National/Tenth Conference on
  Artificial Intelligence/Innovative Applications of Artificial Intelligence},
  AAAI '98/IAAI '98, page 425–430, USA, 1998. American Association for
  Artificial Intelligence.
\newblock ISBN 0262510987.

\bibitem[Zohar and Wolf(2018)]{zohar2018automatic}
Amit Zohar and Lior Wolf.
\newblock Automatic program synthesis of long programs with a learned garbage
  collector.
\newblock In \emph{Advances in Neural Information Processing Systems}, pages
  2094--2103, 2018.

\end{thebibliography}
}

\newpage
\appendix

\section{Details of PCCoder~\cite{zohar2018automatic}}
\label{app:pccoder}
We provide our version of the training and inference algorithms and description of modules used in PCCoder~\cite{zohar2018automatic} next. Note that the terminology used in PCCoder differs from what we have used here.

\subsection{Training and Inference Algorithms of PCCoder}
\label{app:pccoder_algos}
\begin{minipage}{0.46\textwidth}
\begin{algorithm}[H]
    \scriptsize
    \centering
    \caption{Train (GPS)}
    \label{alg:pccoder_train}
    \begin{algorithmic}[1]
    \Require $p_g = [p_g^t]_{t=1}^{T}$ = ground-truth program with $T$ lines
    \Require $\nu$ = max \# allowed variables = memory-size
    \Require $X = \{(x_i, y_i)\}_{i=1}^{N} = \{r_i\}_{i=1}^{N}$ = set of $N$ IO examples
    \Statex
    
    \State $\mathcal{X}^0 = [r_i]_{i=1}^n$  \Comment{\textcolor{blue}{\textit{Initial State}}}
    \For{$t$ in $range(T))$} 
    \LineComment{\textcolor{blue}{\textit{Obtain ground truth}}}
    \State $s^t, o^t, d^t$ = $R(p_g^t, [p_g^j]_{j \geq t}, \mathcal{H}^{t-1})$ 
    \State $\mathcal{H}^{t-1} = H_{\theta}(\mathcal{X}^{t-1}$) \Comment{\textcolor{blue}{\textit{Obtain current state embedding}}}
    \State $\hat{s}^t, \hat{o}^t, \hat{d}^t = W_{\phi}(\mathcal{H}^{t-1})$ \Comment{\textcolor{blue}{\textit{Obtain predictions}}}
    \LineComment{\textcolor{blue}{\textit{Calculate loss and update parameters}}}
    \State $\mathcal{L}$ = CE ($s^t, \hat{s}^t$ ) + CE ($o^t, \hat{o}^t$) + $\Sigma_{j=1}^{\nu}$ BCE ($d^{t^{j}}, \hat{d}^{t^{j}})$
    \State $\theta \leftarrow \theta - \alpha * \nabla_{\theta}\mathcal{L}$
    \State $\phi \leftarrow \phi - \alpha * \nabla_{\phi}\mathcal{L}$
    \LineComment{\textcolor{blue}{\textit{Randomly chose an index to drop}}}
    \State $d^{\prime^{t}}$ = random\_choice($d^t$)
    \LineComment{\textcolor{blue}{\textit{Execute $p_g^t$ to get updated state}}}
    \State $\mathcal{X}^t = DropExec(p_g^t, \mathcal{X}^{t-1}, d^{\prime^{t}}, \nu)$
    \EndFor
    \Statex
    \hrule
    \Procedure{\textcolor{red}{DropExec}} {$p, x, d^\prime, \nu$}

    \State $l = get\_num\_vars(x)$ 
    \State $N = shape(x)[0]$ \Comment{\textcolor{blue}{\textit{\# IO examples}}}
    \For{$i$ in range($N$)}
    \State $x_i = x[i]$
    \LineComment{\textcolor{blue}{\textit{Execute $p$ against $x_i$ to obtain result $c_i$}}}
    \State $c_i = Execute (p, x_i)$ 
    \If{$l > \nu$} \Comment{\textcolor{blue}{\textit{Need to drop a variable}}}
    \State $x_i[d^\prime] = c_i$
    \Else 
    \State $x_i.append(c_i)$
    \EndIf
    \EndFor
    \State $l = l+1$
    \State $set\_num\_vars(x, l)$
    \State \textbf{return} $x$ \Comment{\textcolor{blue}{\textit{return the updated state}}}
    \EndProcedure
    \end{algorithmic}
    \end{algorithm}
    \end{minipage}
\hfill
\begin{minipage}{0.46\textwidth}
\begin{algorithm}[H]
    \scriptsize
    \centering
    \caption{Inference (GPS)}
    \label{alg:pccoder_infer}
    \begin{algorithmic}[1]
    \State $\mathcal{X}^0 = [r_i]_{i=1}^n$  
    \While{time < timeout} \Comment{CAB outer loop}
    \LineComment{\textcolor{blue}{\textit{Initial Beam: (state, program, prob)}}}
    \State $B = [(\mathcal{X}^0, [ \hspace{0.2cm} ], 1.0)]$
    \LineComment{\textcolor{blue}{\textit{$p_g=$ global solution}}}
    \State $p_g = BeamSearch(B)$    \Comment{CAB inner loop}
    \If{$p_g$ == FAILED}
    \State beam\_size*=2; beam\_expansion\_size+=10
    \EndIf
    \EndWhile
    \Statex
    \hrule
    \Procedure{\textcolor{red}{BeamSearch}} {$B$}

    \While{beam search conditions are met} 
    \State $B^\prime$ = [\hspace{0.2cm}] \Comment{\textcolor{blue}{\textit{new beams}}}
    \LineComment{\textcolor{blue}{\textit{For each parent node}}}
    \For{($b, (\mathcal{X}^{t-1}_b, p^{t-1}_b, s^{t-1}_b))$ in enum($B$)} 
    \If{is\_solution($\mathcal{X}^{t-1}_b$)}
    \State \textbf{return} $p^{t-1}_b$
    \EndIf
    \State $\mathcal{H}^{t-1}_b = H_{\theta}(\mathcal{X}^{t-1}_b)$
    \State $\hat{s}^t_b, _ , \hat{d}^t_b = W_{\phi} (\mathcal{H}^{t-1}_b)$
    \LineComment{\textcolor{blue}{\textit{sort $\hat{s}^t_b$ by decreasing probability}}}
    \State $\hat{s}^t_b = sort (\hat{s}^t_b) $
    \LineComment{\textcolor{blue}{\textit{choose argmax of $\hat{d}^t_b$ to drop }}}
    \State $d^{\prime^{t}}_b = argmax(\hat{d}^t_b)$
    \LineComment{\textcolor{blue}{\textit{Expand the parent node}}}
    \For{$\tilde{s}^t_b$ in $\hat{s}^t_b[: expansion\_size]$}
    \LineComment{\textcolor{blue}{\textit{get statement id for the prob entry}}}
    \State $p^t_b = prob\_to\_stat(\tilde{s}^t_b)$ 
    \LineComment{\textcolor{blue}{\textit{get updated memory}}}
    \State $\mathcal{X}_b^{t} = DropExec(p^t_b, \mathcal{X}^{t-1}_b, d^{\prime^{t}}_b, \nu)$
    \State $p^{t-1}_b.append(p^t_b)$ \Comment{\textcolor{blue}{\textit{updated program}}}
    \State $s^{t-1}_b = s^{t-1}_b* \tilde{s}^t_b$ \Comment{\textcolor{blue}{\textit{updated probability}}}
    \State $B^\prime$.append(($\mathcal{X}^t_b, p^{t-1}_b, s^{t-1}_b$))
    \EndFor
    \EndFor
    \LineComment{\textcolor{blue}{\textit{sort beams by decreasing probability}}}
    \State $B^\prime$ = sort ($B^\prime$) [:beam\_size] 
    \State $B$ = $B^\prime$
    \EndWhile
    \State \textbf{return} FAILED \Comment{\textcolor{blue}{\textit{if no solution found during beam search, return Failed solution}}}
    \EndProcedure
    \end{algorithmic}
    \end{algorithm}
    \end{minipage}

\subsection{Description of Modules in PCCoder}
\label{app:state}

The inputs to a program can either be an integer or an array of integers of maximum length 20. The integers can be in the range [-256, 255]. There can be a maximum of three input arguments to a program. There are 1298 statements and 38 operators in the DSL, i.e., $n_s=1298$ and $n_o=38$. Execution of a line in the program returns exactly one variable. Below, we describe the blocks present at different stages of PCCoder:
\begin{itemize}

    \item \textbf{State Representation:} For each of the $N$ IO examples, the corresponding inputs are taken and all entries are made positive by subtracting the minimum integer value under the DSL (i.e.\ -256) from them. For shorter inputs, NULL values up to length $q=20$ are padded. Then two bits indicating the type of input (list or int) are appended at the beginning of this representation. Therefore, each variable is now represented as a vector of size $q$ + 2 = 22. There can be a maximum of $\nu$ input variables and one output variable (corresponding to the output of the IO example given). If there are less than $\nu$ variables, NULL values are padded to make it uniform. An account of the actual number of variables (i.e. number of filled slots) present in the state is also kept, denoted by $l$. The output of this stage is an array of size $N \times (\nu +1) \times (q+2)$. This forms the \emph{state} $\mathcal{X}$.
    
    \item \textbf{State Embedding ($\mathbf{H_{\theta}}$):} The output obtained in the previous step is then passed through a series of neural network blocks to obtain a \emph{state embedding} $\mathcal{H}$. An embedding layer projects each entry in the state (excluding the type bits) into an vector of size $e=20$, giving us a tensor of size $N \times (\nu +1) \times (q*e+2)$. This is then passed through a linear layer of size 56 and then reshaped to obtain a tensor of size $N \times (\nu +1) * 56$. It is then passed through a dense block to obtain a tensor of size $N \times Z$ where $Z=256$. This pre-pooling version is what we refer to as the representation of slots in Section 3.2. An average pooling of these representations across all $N$ examples gives a vector of size $1 \times 256$ that forms the state embedding.
    
    \item \textbf{Predicting quantities of interest in next line ($\mathbf{W_{\phi}}$):} The state embedding obtained above is projected into three linear heads of size 1298, 38 and $\nu$ followed by softmax, softmax and sigmoid, respectively which gets us the statement, operator and drop probabilities. 

    \item \textbf{DropExec:} In the $DropExec$ module, after a statement is executed against the variables present in the slots in the state $\mathcal{X}^0$, we get new values of resulting variables. If the actual number of variables $l$ exceeds $\nu$, one of the existing variables is dropped based on the drop vector. If not, this new variable is simply appended to the existing variables by filling the next slot in the memory. This updated state is then passed through $H_{\theta}$ to get the updated state embedding $\mathcal{H}^1$. This completes one step of execution of the program. 
\end{itemize}

For the next steps, we repeat the last two steps mentioned above till we reach the end of the program. See Figure 2 for an illustration of the process at $t=2$.

\section{Sample Cases}
\label{app:sample_cases}
Below we provide two sample cases where GPS fails and our N-PEPS model (for E2) succeeds in finding a global solution. Foe each sample case, we show the synthesized global solution on the left, the set of IO examples in the center and the discovered PE solutions along with PE solution scores in the right. We also report the actual time taken to find the solutions. Note that for the second case, even though the global ground-truth test program is of length 8, N-PEPS discovers a global solution of shorter length.\\

\fbox{
\begin{minipage}{.31\textwidth}
	\textbf{Global Solution:}\\
	(Time taken to find=3.21s) \\ \\
    \texttt{a} $\gets$ \texttt{LIST}\\
    \texttt{b} $\gets$ \textsc{ZipWith} \texttt{(+)} \texttt{a} \texttt{a}\\
    \texttt{c} $\gets$ \textsc{Tail} \texttt{b}\\
    \texttt{d} $\gets$ \textsc{Take} \texttt{c} \texttt{b}\\
    \texttt{e} $\gets$ \textsc{Count} \texttt{(>0)} \texttt{d}\\
    \texttt{f} $\gets$ \textsc{Take} \texttt{e} \texttt{d}\\
    \texttt{g} $\gets$ \textsc{Count} \texttt{(>0)} \texttt{f}\\
    \texttt{h} $\gets$ \textsc{Take} \texttt{g} \texttt{f}\\
    \texttt{i} $\gets$ \textsc{Take} \texttt{g} \texttt{h}\\
    \texttt{j} $\gets$ \textsc{Head} \texttt{i}\\
    \texttt{k} $\gets$ \textsc{Take} \texttt{j} \texttt{i}\\
    \texttt{l} $\gets$ \textsc{Take} \texttt{j} \texttt{k}\\
    \texttt{m} $\gets$ \textsc{Take} \texttt{j} \texttt{k}\\
    \texttt{n} $\gets$ \textsc{Take} \texttt{j} \texttt{k}\\
    \texttt{o} $\gets$ \textsc{Reverse} \texttt{n}\\
    \end{minipage}
    \begin{minipage}{.35\textwidth}
    \textbf{IO examples:}\\
    \textbf{\#1.} \emph{Input}: \\
    \texttt{[4, 5, 6, 2, 6, 2, 1, 6, 1, 4, 2, 5, 6, 3, 2, 2]}\\
    \emph{Output}:\\
    \texttt{[4, 12, 10, 8]}\\
    \textbf{\#2.} \emph{Input}: \\
    \texttt{[3, 2, 5, 0, 3, 2, 3, 0, 4, 1, 0, 2, 3, 0, 3, 4]}\\
    \emph{Output}:\\
    \texttt{[6, 0, 10, 4, 6]}\\
    \textbf{\#3.} \emph{Input}: \\
    \texttt{[1, 1, 4, 0, 0, 0, 0, 5, 0, 5, 3, 5]}\\
    \emph{Output}:\\
    \texttt{[2, 2]}\\
    \textbf{\#4.} \emph{Input}: \\
    \texttt{[4, 4, 1, 4, 4, 1, 4, 2, 2, 1, 3, 4]}\\
    \emph{Output}:\\
    \texttt{[4, 8, 2, 8, 8, 2, 8, 8]}\\
    \textbf{\#5.} \emph{Input}: \\
    \texttt{[4, 1, 1,, 3, 3, 1, 4, 0, 4, 2, 4]}\\
    \emph{Output}:\\
    \texttt{[8, 2, 6, 6, 2, 2, 8]}\\
    \end{minipage} \qquad
    \hfill
    \begin{minipage}{.31\textwidth}
    \textbf{PE Solutions:}\\
    $\mathbf{p_1}:$ Time taken to find=0.2s\\
    Satisfies \#1, \#4 ($u_1=0.2$)\\\\
    \texttt{a} $\gets$ \texttt{LIST}\\
    \texttt{b} $\gets$ \textsc{ZipWith} \texttt{(+)} \texttt{a} \texttt{a}\\
    \texttt{c} $\gets$ \textsc{Tail} \texttt{b}\\
    \texttt{d} $\gets$ \textsc{Take} \texttt{c} \texttt{b}\\
    \texttt{e} $\gets$ \textsc{Reverse} \texttt{d}\\ \\
    
    $\mathbf{p_2}:$ Time taken to find=0.34s\\
    Satisfies \#2,  \#3, \#4, \#5 ($u_2=0.8$)\\\\
    \texttt{a} $\gets$ \texttt{LIST}\\
    \texttt{b} $\gets$ \textsc{ZipWith} \texttt{(+)} \texttt{a} \texttt{a}\\
    \texttt{c} $\gets$ \textsc{Head} \texttt{b}\\
    \texttt{d} $\gets$ \textsc{Take} \texttt{c} \texttt{b}\\
    \texttt{e} $\gets$ \textsc{Count} \texttt{(>0)} \texttt{d}\\
    \texttt{f} $\gets$ \textsc{Take} \texttt{e} \texttt{d}\\
    \texttt{g} $\gets$ \textsc{Reverse} \texttt{f}\\
    \end{minipage}
}

\fbox{
\begin{minipage}{.31\textwidth}
	\textbf{Global Solution:}\\
	(Time taken to find=2.98s) \\ \\
    \texttt{a} $\gets$ \texttt{LIST}\\
    \texttt{b} $\gets$ \texttt{INT}\\
    \texttt{c} $\gets$ \textsc{Maximum} \texttt{a}\\
    \texttt{d} $\gets$ \textsc{Take} \texttt{c} \texttt{a}\\
    \texttt{e} $\gets$ \textsc{Tail} \texttt{c}\\
    \texttt{f} $\gets$ \textsc{Take} \texttt{b} \texttt{c}\\
    \texttt{g} $\gets$ \textsc{ZipWith} \texttt{(+)} \texttt{f} \texttt{f}\\
    \texttt{h} $\gets$ \textsc{Map} \texttt{(+1)} \texttt{g}\\
    \texttt{i} $\gets$ \textsc{Take} \texttt{e} \texttt{h}\\
    \end{minipage}
    \begin{minipage}{.35\textwidth}
    \textbf{IO examples:}\\
    \textbf{\#1.} \emph{Input}: \\
    \texttt{[1, 0, 3, 3, 3], 35}\\
    \emph{Output}:\\
    \texttt{[3, 1, 7]}\\
    \textbf{\#2.} \emph{Input}: \\
    \texttt{[6, 3, 3, 1, 2, 2, 0, 3, 8, 7], 50}\\
    \emph{Output}:\\
    \texttt{[13, 7, 7]}\\
    \textbf{\#3.} \emph{Input}: \\
    \texttt{[1, 5, 6, 10, 5, 11, 7, 0, 7, 11, 10, 9, 4], 78}\\
    \emph{Output}:\\
    \texttt{[3, 11, 13, 21, 11, 23, 15, 1, 15, 23]}\\
    \textbf{\#4.} \emph{Input}: \\
    \texttt{[12, 4, 11, 11, 4, 7, 12, 11, 11, 10, 5, 8, 9, 8], 166}\\
    \emph{Output}:\\
    \texttt{[25, 9, 23, 23, 9, 15, 25, 23]}\\
    \textbf{\#5.} \emph{Input}: \\
    \texttt{[4, 0, 5, 5, 1, 1, 1, 1], 126}\\
    \emph{Output}:\\
    \texttt{[9]}\\
    \end{minipage} \qquad
    \hfill
    \begin{minipage}{.31\textwidth}
    \textbf{PE Solutions:}\\
    $\mathbf{p_1}:$ Time taken to find=0.17s\\
    Satisfies \#1, \#4, \#5 ($u_1=0.6$)\\\\
    \texttt{a} $\gets$ \texttt{LIST}\\
    \texttt{b} $\gets$ \texttt{INT}\\
    \texttt{c} $\gets$ \textsc{Tail} \texttt{a}\\
    \texttt{d} $\gets$ \textsc{Take} \texttt{c} \texttt{a}\\
    \texttt{e} $\gets$ \textsc{ZipWith} \texttt{(+)} \texttt{d} \texttt{d}\\
    \texttt{f} $\gets$ \textsc{Map} \texttt{(+1)} \texttt{e}\\ \\

    $\mathbf{p_2}:$ Time taken to find=0.37s\\
    Satisfies \#1, \#5 ($u_2=0.4$)\\\\
    \texttt{a} $\gets$ \texttt{LIST}\\
    \texttt{b} $\gets$ \texttt{INT}\\
    \texttt{c} $\gets$ \textsc{Take} \texttt{b} \texttt{a}\\
    \texttt{d} $\gets$ \textsc{Tail} \texttt{c}\\
    \texttt{e} $\gets$ \textsc{Access} \texttt{d} \texttt{c}\\
    \texttt{f} $\gets$ \textsc{Take} \texttt{e} \texttt{c}\\
    \texttt{g} $\gets$ \textsc{ZipWith} \texttt{(+)} \texttt{f} \texttt{f}\\
    \texttt{h} $\gets$ \textsc{Map} \texttt{(+1)} \texttt{g}\\ \\
    
    $\mathbf{p_3}:$ Time taken to find=0.8s\\
    Satisfies None ($u_3=0.0$)\\ \\
    \texttt{FAILED} \\ \\
    
    $\mathbf{p_4}:$ Time taken to find=0.17s\\
    Satisfies \#1, \#4, \#5 ($u_4=0.6$)\\\\
    \texttt{a} $\gets$ \texttt{LIST}\\
    \texttt{b} $\gets$ \texttt{INT}\\
    \texttt{c} $\gets$ \textsc{Tail} \texttt{a}\\
    \texttt{d} $\gets$ \textsc{Take} \texttt{c} \texttt{a}\\
    \texttt{e} $\gets$ \textsc{ZipWith} \texttt{(+)} \texttt{d} \texttt{d}\\
    \texttt{f} $\gets$ \textsc{Map} \texttt{(+1)} \texttt{e}\\ \\
    
    $\mathbf{p_5}:$ Time taken to find=0.17s\\
    Satisfies \#1, \#4, \#5 ($u_5=0.6$)\\\\
    \texttt{a} $\gets$ \texttt{LIST}\\
    \texttt{b} $\gets$ \texttt{INT}\\
    \texttt{c} $\gets$ \textsc{Tail} \texttt{a}\\
    \texttt{d} $\gets$ \textsc{Take} \texttt{c} \texttt{a}\\
    \texttt{e} $\gets$ \textsc{ZipWith} \texttt{(+)} \texttt{d} \texttt{d}\\
    \texttt{f} $\gets$ \textsc{Map} \texttt{(+1)} \texttt{e}\\ \\
    \end{minipage}
}

\section{Data Generation}
\label{app:data_gen}

\subsection{Generation of Training and Test set}
\label{app:train_test_prog_gen}

Similar to the data generation process described in \citet{balog2016deepcoder, zohar2018automatic} and using the implementation from PCCoder~\footnote{https://github.com/amitz25/PCCoder (MIT License)}, we generated programs for training and testing where each program consists of five input-output examples. The process starts by generating training programs iteratively starting from length 1 till the maximum length specified (4 and 12 in our case). For each length, first a program of that length is generated followed by generating corresponding IO examples which correspond to that program. This is followed by checking for functional non-equivalence of that program with all generated programs so far (i.e., programs of length less than or equal to the current length). Functional non-equivalence means that given a set of IO examples, we can't have a program of length x that satisfies the set of examples when we already have a program of length less than or equal to x in our dataset that satisfies the same set of examples. If the program is found functionally equivalent to any other programs, it is discarded, else it is added to the training set. 

\begin{figure}[htb!]
\begin{subfigure}{.48\textwidth}
    \centering
     \includegraphics[width=1.0\linewidth]{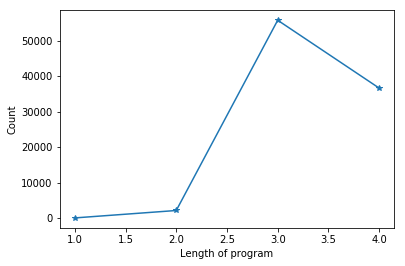}
\end{subfigure}
\hfill
\begin{subfigure}{.48\textwidth}
    \centering
     \includegraphics[width=1.0\linewidth]{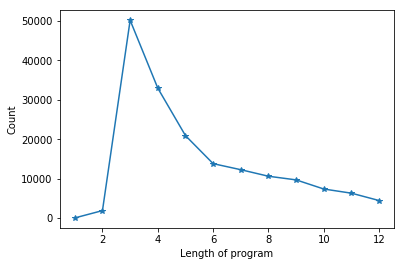}
\end{subfigure}
\caption{\textbf{Distribution of training programs}: \textbf{\textit{(Left)}} For E1; \textbf{\textit{(Right)}} For E2}
\label{fig:data_dist}
\vskip -0.15in
\end{figure}

Once the generation of training set is complete, we proceed to generating the test set. Given a test length, we generate a program of that length followed by generating the corresponding IO example pair. In addition to checking for functional non-equivalence with all programs in the test set so far, we also test for functional non-equivalence with every program in the training set. This makes sure that there there is no overlap between the training and test sets and all the programs are functionally non-equivalent to each other. We have two experimental settings: (a) \textbf{E1:} Training set = 105036 programs till length 4 and 30 test sets of 500 programs each of length = 4; (b) \textbf{E2:} Training set = 189328 programs of length up to 12 and 30 test sets of 500 programs each of lengths = 5, 8, 10, 12 and 14. In each setting, 10\% of the training data was used for validation. Figure~\ref{fig:data_dist} shows the distribution of training programs with length in both the settings. There are less programs of longer lengths as there is high probability that they end up being discarded because a functionally equivalent program of shorter length was found.

\subsection{Generation of Aggregator Instances}
\label{app:agg_ins_gen}

An aggregator instance consists of the set of IO examples $X$, a list $Y$ of PE solutions $p_i$ along with the corresponding PE solution scores $u_i$, and the corresponding global program $p_g$. To create aggregator instances, for each data point (given $X$ and $p_g$) in the original training dataset (generated as described in \ref{app:train_test_prog_gen}), we generate PE solutions and PE solution scores using the PE Searches module. For generating the PE solutions, we need to choose a value of PEPS timeout. We generated aggregator instances with PE solutions obtained using the trained PE model, in three ways: (a) one aggregator instance with a fixed PEPS timeout of 0.5s; (b) two aggregator instances with PEPS timeout randomly chosen from [0.1s, 0.2s, .. 0.9s]; (c) three aggregator instances each with PEPS timeouts of 0.4s, 0.5s and 0.6s, respectively. These options will lead to the same, twice and thrice the number of data points present in the original training set. We chose to omit a sample from being part of training data formed from aggregator instances if either (a) An aggregator instance consists of a PE solution that satisfies all examples (i.e., $u_i = 1.0$) or (b) When we fail to get any PE solution (i.e., all $u_i = 0.0$). We can then generate data which omits both (a) and (b). The datasets formed after removing these aggregator instances will be referred to as
$\mathbf{D_{0.5}}$, $\mathbf{D_{rand}}$ and $\mathbf{D_{0.5 \pm 0.1}}$ for cases (a), (b) and (c), respectively. In addition, within each aggregator instance, we can chose to discover all 5 PE solutions (\textbf{\emph{all}}) or alternatively find a list of $M$ PE solutions where $M \leq 5$ such that taken together these satisfy all examples in $X$ (\textbf{\emph{tot}}). Therefore, in total we have 12 different variations (3 PEPS timeouts $\times$ 2 inclusion conditions $\times$ 2 modes of discovering PE solutions) of training datasets which can be used for training the cross-aggregator. We follow a similar procedure to generate variations of corresponding validation datasets which are used to select the hyperparameters and early-stopping for training with each dataset variation. Table~\ref{tab:agg_data_stats} gives the training and validation data statistics (note that the 2 modes for discovering PE solutions will affect the content of a single aggregator instance, but the number of aggregator instances will remain the same in both cases). 

\begin{table}[t]
\small
\centering
\caption{Aggregator data statistics for E1 and E2}
\vskip 0.2cm
\begin{tabular}{ccc}
	    \toprule
    	Dataset & \multicolumn{2}{c}{\# of samples}\\
    	\cline{2-3}\\
    	& E1 & E2\\
        \midrule
        \vspace{0.15cm}
        $D_{0.5}^{train}$ & 16408 & 82102 \\ 
        \vspace{0.15cm}
        $D_{0.5}^{val}$ & 2132 & 9492\\ 
        \vspace{0.15cm}
        $D_{rand}^{train}$ & 41707 & 176235\\
        \vspace{0.15cm}
		$D_{rand}^{val}$ & 5365 & 20035\\ 
		\vspace{0.15cm}
		$D_{0.5 \pm 0.1}^{train}$ & 49116 & 248972\\ 
		\vspace{0.15cm}
		$D_{0.5 \pm 0.1}^{val}$ & 6396 & 28734\\ 
		\bottomrule
	\end{tabular}
\label{tab:agg_data_stats}
\end{table}

\section{Experimental details}
\label{app:hyperparams}
All our implementations are based on PyTorch~\cite{NEURIPS2019_9015} version 1.4.0. The training for the GPS and PE models and the CA was done using Tesla V100 (16GB) and Tesla P100 (16GB) GPUs on Google Cloud instances with 120GB RAM.

\subsection{Parallel Execution for PEPS}
In the current formulation, we find PE solutions sequentially. However, the running time can be reduced further by finding PE solution in parallel as the process of finding PE solution $i$ is dependent only on the IO example $r_i$. So, instead of finding PE solutions one by one, we can find PE solutions for all examples in parallel and then check whether the PE solution $p_i$ satisfy any other example from $X$ apart from $r_i$. The total time for PEPS can then be thought of max(time taken to find a PE solution that satisfies $r_i$) + time taken to aggregate. However, one could argue that PCCoder can also employ more threads in parallel to speed up their search. Therefore, for a fair comparison with PCCoder, we decided to find PE solutions sequentially where we evaluate both N-PEPS and PCCoder on a single CPU thread (with no parallel computations). However, when being deployed for an application where comparisons with other methods are not required, N-PEPS can significantly boost the speed up by searching for PE solutions in parallel in a way suggested above.

\subsection{Training the GPS and PE models}
For each experimental setting, we used the training set (generated as described in \ref{app:train_test_prog_gen}) as it is for training the GPS model. For training the PE model, we created five entries out of a single training data point such that a modified entry has a single IO example and the corresponding program is the same across all five entries = program in the data point for GPS. Since we don't have supervision available for PE solutions, we chose $p_g$ to serve as a proxy for ground-truth of these PE solutions. Another way of creating this supervision would have been to perform separate PE searches for each example and recording the discovered PE solution as ground-truth. However, this procedure would have required the selection of a specific PE timeout. We didn’t have any good idea of how to select this value as it would have influenced the generated PE solutions, hence the supervision itself. Also, we didn’t know what would have been the best supervision to use for cases where the PE search fails to find a solution. We believe that using $p_g$ as proxy supervision even though not being entirely correct, forces the PE search component to avoid overfitting to a single example and hence is more likely to produce PE solutions that exhibit intent generalization (generalization to examples outside the ones given as specification). 

The number of training points for PE model = 5 * number of training data points for GPS.  The corresponding validation split was used to select hyperparameters. The selected hyperparameter values were:
\begin{itemize}
    \item \emph{GPS model:} learning rate = 0.001; batch size = 32 for E1 and 100 for E2.
    \item \emph{PE model:} learning rate = 0.001; batch size = 100 for E1 and 256 for E2.
\end{itemize}
For both settings $\nu=11$. This means that the state has slot for storing 7 intermediate variables, 3 slots for input variables (there can be a maximum of 3 input arguments to a program) and an additional slot for storing the output. This means that for E2, dropping will happen for programs of length greater than 8. We used Adam~\cite{DBLP:journals/corr/KingmaB14} optimizer with a learning rate scheduler that decayed the learning rate by 0.1 every 4 epochs. We used the validation set for early-stopping. Let's call the learned PE modules as $H_{\theta_{pe}}$ and $W_{\phi_{pe}}$ whereas, the corresponding GPS modules to be $H_{\theta_{g}}$ and $W_{\phi_{g}}$.

\subsection{Training the Cross Aggregator}
\label{app:train_ca}
For both E1 and E2, we train our cross aggregator (CA) module using the variants of keys mentioned in Section 3.2 and Section 4.3. For N-PEPS-PG, we use $H_{\theta_{g}}$ to obtain state embeddings that forms the keys, whereas for N-PEPS and N-PEPS-PP we used $H_{\theta_{pe}}$. For faster convergence, we initialize the statement and operator heads with the corresponding statement and operator linear heads from $W_{\phi_{g}}$. We tried finetuning the parameters of $H$, but it didn't result in significant difference in training performance. Hence, we decided to leave the parameters of the $H$ module unaltered during training. As mentioned in \ref{app:agg_ins_gen}, we tried both \emph{all} and \emph{tot} ways of discovering PE solutions while training. In equations 1, 2 and 3 in Section 3.2, the projection matrices $W_i^{Q} \in \mathbb{R}^{d_{model} \times d_q}$, $W_i^{K} \in \mathbb{R}^{d_{model} \times d_k}$, $W_i^{V} \in \mathbb{R}^{d_{model} \times d_v}$, $W^{O} \in \mathbb{R}^{\tau d_v \times d_{model}}$. For the multihead relative attention, we used $d_k = d_q = d_v = 64$, $\tau=8$ and $d_{model}= 256$. A dropout value of 0.1 was used while training.

\subsection{Details of Training Hyperparameters}
We tried different values of learning rates, optimizers, learning rate schedulers, datasets and the PE discovery options. 
We tried three types of learning rate schedulers\footnote{see https://pytorch.org/docs/stable/optim.html for more details}: (a) \textbf{cosine}: \code{torch.optim.lr\_scheduler.CosineAnnealingLR(optimizer, T\_max=10, eta\_min=0)}; (b) \textbf{cosinewarm}: \code{torch.optim.lr\_scheduler.CosineAnnealingWarmRestarts(optimizer, T\_0=10)} ; (c) \textbf{reduceonplateau}: \code{torch.optim.lr\_scheduler.ReduceLROnPlateau(optimizer, 'min')} where \code{optimizer = Adam, SGD}.
Below we provide the hyperparameter configuration for the best models chosen using the validation set.

\begin{table}[htb!]
\scriptsize
\centering
\caption{Hyperparameter values for training the CA. lr= learning rate, lrs = learning rate scheduler, o=optimizer}
\vskip 0.2cm
\begin{tabular}{ccc}
	    \toprule
    	Model & \multicolumn{2}{c}{Hyperparameters}\\
    	\cline{2-3}\\
    	& E1 & E2\\
        \midrule
        \vspace{0.15cm}
        N-PEPS-PP & $D_{0.5}$, \emph{all}, lr=1e-4, o =Adam, lrs=cosine &  $D_{0.5 \pm 0.1}$, \emph{tot}, lr=1e-4, o=Adam, lrs=reduceonplateau\\
        \vspace{0.15cm}
        N-PEPS-PP$+\mathcal{U}$ & $D_{rand}$, \emph{all}, lr=1e-4, o=SGD, lrs=cosinewarm  &  $D_{rand}$, \emph{all}, lr=1e-4, o=SGD, lrs=cosinewarm\\
        \vspace{0.15cm}
        N-PEPS-PG & $D_{rand}$, \emph{tot}, lr=1e-4, o=SGD, lrs=cosine  &  $D_{0.5}$, \emph{all}, lr=1e-4, o=SGD, lrs=cosine \\
        \vspace{0.15cm}
        N-PEPS-PG$+\mathcal{U}$ & $D_{rand}$, \emph{all}, lr=1e-4, o=SGD, lrs=cosinewarm  &  $D_{0.5}$, \emph{all}, lr=1e-4, o=Adam, lrs=reduceonplateau\\
         \vspace{0.15cm}
        N-PEPS & $D_{rand}$, \emph{all}, lr=1e-4, o=SGD, lrs=cosine  &  $D_{rand}$, \emph{all}, lr=3e-4, o=Adam, lrs=cosine\\
        \vspace{0.15cm}
        N-PEPS$+\mathcal{U}$ & $D_{0.5}$, \emph{tot}, lr=3e-4, o=Adam, lrs=cosinewarm  &  $D_{rand}$, \emph{all}, lr=1e-4, o=Adam, lrs=reduceonplateau\\
		\bottomrule
	\end{tabular}
\label{tab:ca_hyperparams}
\end{table}

\subsection{Details of Inference Hyperparameters}

For inference we use CAB~\cite{cab} which consists of performing beam search iteratively, with pruning conditions of beam search (i.e., beam size, expansion size, etc.) weakened with each iteration, until a solution is found. Simialr to PCCoder~\cite{zohar2018automatic}, we start with beam size = 100, expansion size = 10 and maximum depth of beam search = number of steps = maximum program length. If the beam search fails, we double the beam size and increase the expansion size by 10, and perform beam search again with the modified parameters. The beam search terminates if we exceed the timeout. If no solution is found at the end of CAB, we mark that solution as FAILED.

We created a smaller validation split called \emph{smallval} which consists of 5\% of the samples chosen randomly from the larger validation data. The size of smallval is 525 samples and 946 samples for E1 and E2, respectively. We used this set to find optimal
values of PEPS timeout and $\alpha$ for each model. Table~\ref{tab:inf_hyperparams} provides the selected hyperparameter values for all the models in both the settings. Timeout of 5s is divided between PE Searches module and the aggregation + GPS module. The time allocated to latter is denoted by GT in the table. For GPS, since no PE solutions are discovered, the whole timeout is allocated to the GPS inference block and no aggregation happens, i.e., $\alpha = 0.0$.

\begin{table}[htb!]
\small
\centering
\caption{Hyperparameter values for Inference. PT = PEPS timeout, GT = 5 - ( 5 * PT ).}
\vskip 0.2cm
\begin{tabular}{ccc}
	    \toprule
    	Model & \multicolumn{2}{c}{Hyperparameters}\\
    	\cline{2-3}\\
    	& E1 & E2\\
        \midrule
        \vspace{0.15cm}
        GPS & GT=5.0s, PT=0.0s, $\alpha$=0.0 & GT=5.0s, PT=0s, $\alpha$=0.0\\
        \vspace{0.15cm}
        Sum & GT=2.5s, PT=0.5s, $\alpha$=0.8 & GT=0.5s, PT=0.9s, $\alpha$=0.2\\
        \vspace{0.15cm}
        Mean & GT=2.5s, PT=0.5s, $\alpha$=0.8 & GT=0.5s, PT=0.9s, $\alpha$=0.2\\
        \vspace{0.15cm}
        Mean$+\mathcal{U}$ & GT=2.5s, PT=0.5s, $\alpha$=0.9 & GT=0.5s, PT=0.9s, $\alpha$=0.4\\
        \vspace{0.15cm}
        N-PEPS-PP & GT=1.0s, PT=0.8s, $\alpha$=0.8 & GT=1.5s, PT=0.7s, $\alpha$=0.8 \\
        \vspace{0.15cm}
        N-PEPS-PP$+\mathcal{U}$ & GT=1.0s, PT=0.8s, $\alpha$=0.7 & GT=2.5s, PT=0.5s, $\alpha$=1.0\\
        \vspace{0.15cm}
        N-PEPS-PG & GT=1.0s, PT=0.8s, $\alpha$=0.8 & GT=2.0s, PT=0.6s, $\alpha$=0.9\\
        \vspace{0.15cm}
        N-PEPS-PG$+\mathcal{U}$ & GT=1.0s, PT=0.8s, $\alpha$=0.8 & GT=1.0s, PT=0.8s, $\alpha$=1.0\\
        \vspace{0.15cm}
        N-PEPS & GT=0.5s, PT=0.9s, $\alpha$=0.8 & GT=1.0s, PT=0.8s, $\alpha$=0.8\\
        \vspace{0.15cm}
        N-PEPS$+\mathcal{U}$ & GT=1.0s, PT=0.8s, $\alpha$=0.8 & GT=2.0s, PT=0.6s, $\alpha$=0.9\\
		\bottomrule
	\end{tabular}
\label{tab:inf_hyperparams}
\end{table}

\subsection{Variation across different runs and machines}
\label{app:ins_run_variation}

To ensure robustness and reproducibility of our results, we performed experiments with variations along three dimensions: different runs on the same machine, different machines and different test splits. Table~\ref{tab:ins_run_var} presents the results of variation in success ratio for E1 when run across different test splits, machines and runs across a single machine. Each run consists of a single CPU thread and single core setting on a machine (Google Cloud instance with 120GB RAM). We can see that there is very little variation for runs across the same machine. Hence, for our main experiments we chose to report standard error across different test splits with single runs on machines that are chosen randomly from a pool of 7 Google Cloud instances with same configuration.

\begin{table}[htb!]
\tiny
\centering
\caption{Variation in success ratio for runs across the same machine (run1, run2, run3), different machines (M1, M2, M3) and different test splits (split-1, split-2) for E1}
\vskip 0.2cm
\begin{tabular}{|c|ccc|ccc|ccc|ccc|ccc|ccc|}
            \toprule
            & \multicolumn{9}{|c|}{split-1}  & \multicolumn{9}{|c|}{split-2} \\
            \hline
            & \multicolumn{3}{|c|}{M1} & \multicolumn{3}{|c|}{M2} & \multicolumn{3}{|c|}{M3} & \multicolumn{3}{|c|}{M1} & \multicolumn{3}{|c|}{M2} & \multicolumn{3}{|c|}{M3} \\
            \hline
            & run-1   & run-2  & run-3  & run-1   & run-2  & run-3  & run-1   & run-2   & run-3  & run-1   & run-2  & run-3  & run-1   & run-2  & run-3  & run-1   & run-2   & run-3  \\
            \midrule
GPS         & 77.4    & 77.4   & 78     & 77      & 77.8   & 77.2   & 77.2    & 77.2    & 77.6   & 78.4    & 78.2   & 78.6   & 78      & 78     & 78     & 78.2    & 78      & 78     \\
Sum         & 82.8    & 83.2   & 82.8   & 82.6    & 82.8   & 82.8   & 83.2    & 82.8    & 82.8   & 84.6    & 84.6   & 84.8   & 84.4    & 84.4   & 84.4   & 84.6    & 84.6    & 84.6   \\
Mean        & 82.8    & 82.6   & 82.6   & 82.4    & 82.6   & 82.4   & 82.6    & 82.6    & 82.6   & 85      & 85     & 85     & 84.8    & 84.8   & 85     & 85      & 85      & 84.8   \\
Mean$+\mathcal{U}$    & 82.8    & 82.8   & 82.8   & 82.8    & 82.6   & 82.4   & 82.6    & 82.8    & 82.8   & 85      & 85     & 85     & 84.8    & 85     & 84.8   & 85      & 85      & 85     \\
N-PEPS-PP    & 86.8    & 86.8   & 86.6   & 86.6    & 86.6   & 86.6   & 86.6    & 86.6    & 86.8   & 89.4    & 89.4   & 89.4   & 89.2    & 89.2   & 89.2   & 89.4    & 89.2    & 89.4   \\
N-PEPS-PP$+\mathcal{U}$ & 86.4    & 86.6   & 86.6   & 86.4    & 86.4   & 86.4   & 86.6    & 86.6    & 86.6   & 88.6    & 88.6   & 88.6   & 88.4    & 88.4   & 88.4   & 88.6    & 88.6    & 88.6  \\
N-PEPS-PG     & 86.4    & 86.4   & 86.4   & 86.4    & 86.4   & 86.4   & 86.4    & 86.4    & 86.4   & 87.6    & 87.6   & 87.6   & 87.4    & 87.4   & 87.4   & 87.6    & 87.6    & 87.6   \\
N-PEPS-PG$+\mathcal{U}$ & 87.8    & 88     & 88     & 87.8    & 87.8   & 87.8   & 88      & 88      & 87.8   & 89      & 89     & 89     & 88.8    & 89     & 88.8   & 89      & 89      & 88.8 \\ 
\bottomrule
\end{tabular}
\label{tab:ins_run_var}
\end{table}

\section{Results for Variants of Key for E2}
\label{app:key_var_12}

Table~\ref{tab:key_var_12} presents the test results of ablation studies with different variants of keys for E2. Similar to E1, we see that all the variants perform better than the corresponding values for GPS and the three ablation baselines (see Figure 4 in Section 4.3). We also see that the variant mentioned in Section 3.2 (denoted by N-PEPS in the table) performs the best. Note that even though, these results are on test data, we had chosen the best variant based on the results on the validation data.

\begin{table}[htb!]

    \caption{Success Ratio with standard error for key variants for E2}
    \label{tab:key_var_12}
    \centering
    \scriptsize
	\begin{tabular}{c|ccccc}
	\toprule
	Variant & Length = 5 & Length = 8 & Length = 10 & Length = 12 & Length=14\\
	\midrule

 		N-PEPS-PG & 78.49 $\pm$ 0.35 & 45.92 $\pm$ 0.53 & 31.36 $\pm$ 0.33 & 22.83 $\pm$ 0.33 & 17.15 $\pm$ 0.31\\ 

 		N-PEPS-PG$+\mathcal{U}$ & 78.16 $\pm$ 0.30 & 46.37 $\pm$ 0.57 & 31.88 $\pm$ 0.35 & 23.17 $\pm$ 0.33 & 17.62 $\pm$ 0.30\\ 
		
		N-PEPS-PP & 78.74 $\pm$ 0.32 & 45.9 $\pm$ 0.57 & 31.16 $\pm$ 0.33 & 22.67 $\pm$ 0.32 & 16.91 $\pm$ 0.28\\ 

		N-PEPS-PP$+\mathcal{U}$ & 78.87 $\pm$ 0.35 & 44.87 $\pm$ 0.50 & 30.69 $\pm$ 0.41 & 22.43 $\pm$ 0.36 & 16.59 $\pm$ 0.32\\ 
		
		N-PEPS & 79.18 $\pm$ 0.31 & \textbf{47.23 $\pm$ 0.49} & \textbf{32.3 $\pm$ 0.34} & \textbf{23.34 $\pm$ 0.28} & \textbf{17.35 $\pm$ 0.31}\\ 

		N-PEPS$+\mathcal{U}$ & \textbf{79.19 $\pm$ 0.30} & 46.31 $\pm$ 0.61 & 31.84 $\pm$ 0.36 & 22.71 $\pm$ 0.28 & 16.68 $\pm$ 0.21\\ 
	\bottomrule
	\end{tabular}
\end{table}

\section{Additional Results}
\label{app:add_analysis}

\subsection{Longer Timeout Results}
\label{app:longer_timeout}
We wanted to know whether the performance gains of N-PEPS gets translated to scenarios with a higher computational budget (as opposed to a lower budget of 5s in our setting).  We performed inference with a timeout of 1000s using our previously trained models for GPS and N-PEPS in the E2 setting. For one test split consisting of 500 examples of length=12, the success ratios for GPS and N-PEPS were 54.38\% and 57.14\%, respectively. As expected, when given a higher budget, the numbers for both methods increase. However, N-PEPS still outperforms GPS. Note that here we chose the inference hyperparameters based on an educated guess, i.e., $\alpha = 0.8$, PEPS timeout = 160s and the time given to the CA module = 200s. The test performance of N-PEPS is likely to increase further if the values of these hyperparameters are selected from the validation set. This result provides promising evidence towards the wide applicability of our framework for longer timeout settings. 

\subsection{Intent Generalization Results}
\label{app:intent_gen}
There is an interesting scenario of \emph{intent generalization} where generalization to examples outside of those given as specification is required, in assumption that the additional examples sufficiently define the intent. To see how N-PEPS fares in this setting, we performed experiments where we generated 5 additional IO examples apart from the 5 already present as part of our test data and then evaluated whether the discovered global solutions satisfy the newly generated examples. In Table~\ref{tab:intent_gen} we provide the success ratio with standard error for GPS and N-PEPS across 30 test splits. As can be seen from the results that even though the numbers have reduced from those provided in the tables provided in Figures 3 and 4 of our paper (as expected because the examples are outside of the specification), N-PEPS still outperforms GPS in both E1 and E2 across all lengths.

\begin{table}[H]

    \caption{Success Ratio with standard error for intent generalization experiments}
    \label{tab:intent_gen}
    \centering
    \scriptsize
	\begin{tabular}{c|cccccc}
	\toprule
	Method & Length = 4 (E1) & Length = 5 (E2) & Length = 8 (E2) & Length = 10 (E2) & Length = 12 (E2) & Length = 14 (E2)\\
	\midrule

 		GPS & 75.80 $\pm$ 0.38 & 68.31 $\pm$ 0.38 & 33.87 $\pm$ 0.35 & 18.19 $\pm$ 0.30 & 10.99 $\pm$ 0.26 & 7.48 $\pm$ 0.17\\ 

 		N-PEPS & 84.09 $\pm$ 0.27 & 76.16 $\pm$ 0.32 & 36.33 $\pm$ 0.43 & 21.02 $\pm$ 0.29 & 13.17 $\pm$ 0.25 & 9.17 $\pm$ 0.23\\ 
 		
	\bottomrule
	\end{tabular}
\end{table}

\subsection{Function wise performance}
\label{app:func_perf}

\begin{figure}[H]
  \centering
  \includegraphics[width=1.0\textwidth]{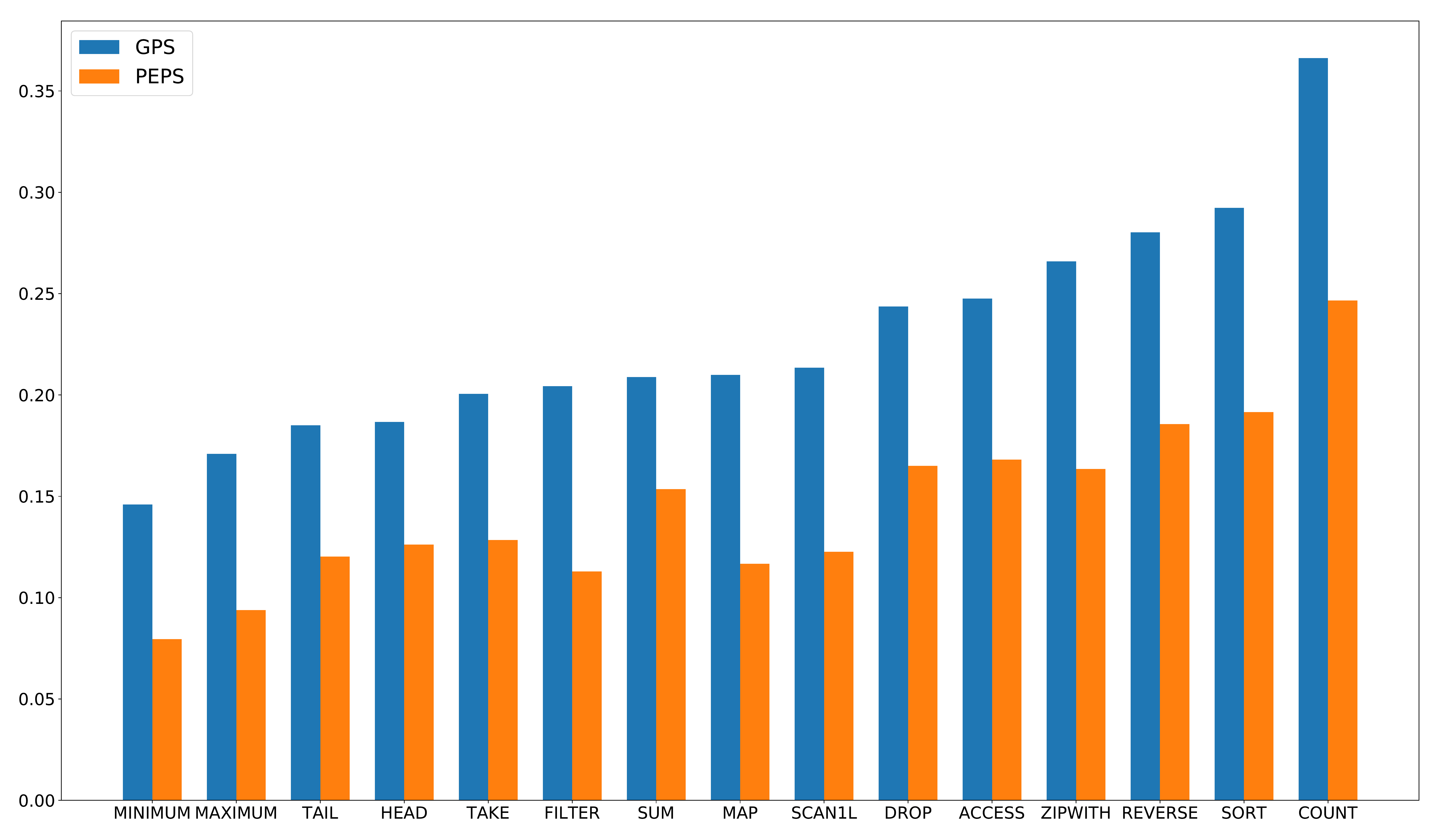}
  \caption{Function-wise breakdown of failing cases for GPS and our N-PEPS model on E1}
  \label{fig:func_len_4}
  \vskip -0.15in
\end{figure}

We wanted to see which instructions in the DSL are "difficult" and compare the difficulty across GPS and N-PEPS. To do this, we record the count of instructions in the cases where the model was not able to find any solution divided by the total count of the instructions. Note that we look only at the operator and not the full statement, i.e., we ignore the arguments. Figure~\ref{fig:func_len_4} shows this plot for GPS and N-PEPS$+\mathcal{U}$ for E1 with numbers across all 30 test splits. We see that usually higher-order functions like \code{COUNT, ZIPWITH} are "difficult" and functions like \code{MAXIMUM, MINIMUM} are comparatively "easy". Also, when compared with GPS, PEPS improves the failure rate across all instructions with improvements ranging from 32.67\% for \code{SUM} to 52.72\% for \code{FILTER}. Other notable improvements being 49.10\% for \code{MAXIMUM}, 44.69\% for \code{MAP}, 45.67\% for \code{SCAN1L} and 47.14\% for \code{TAIL}.

\subsection{Perfect PE solutions}
\label{app:perf_soln}
\begin{figure}[H]
  \centering
  \includegraphics[width=0.7\textwidth]{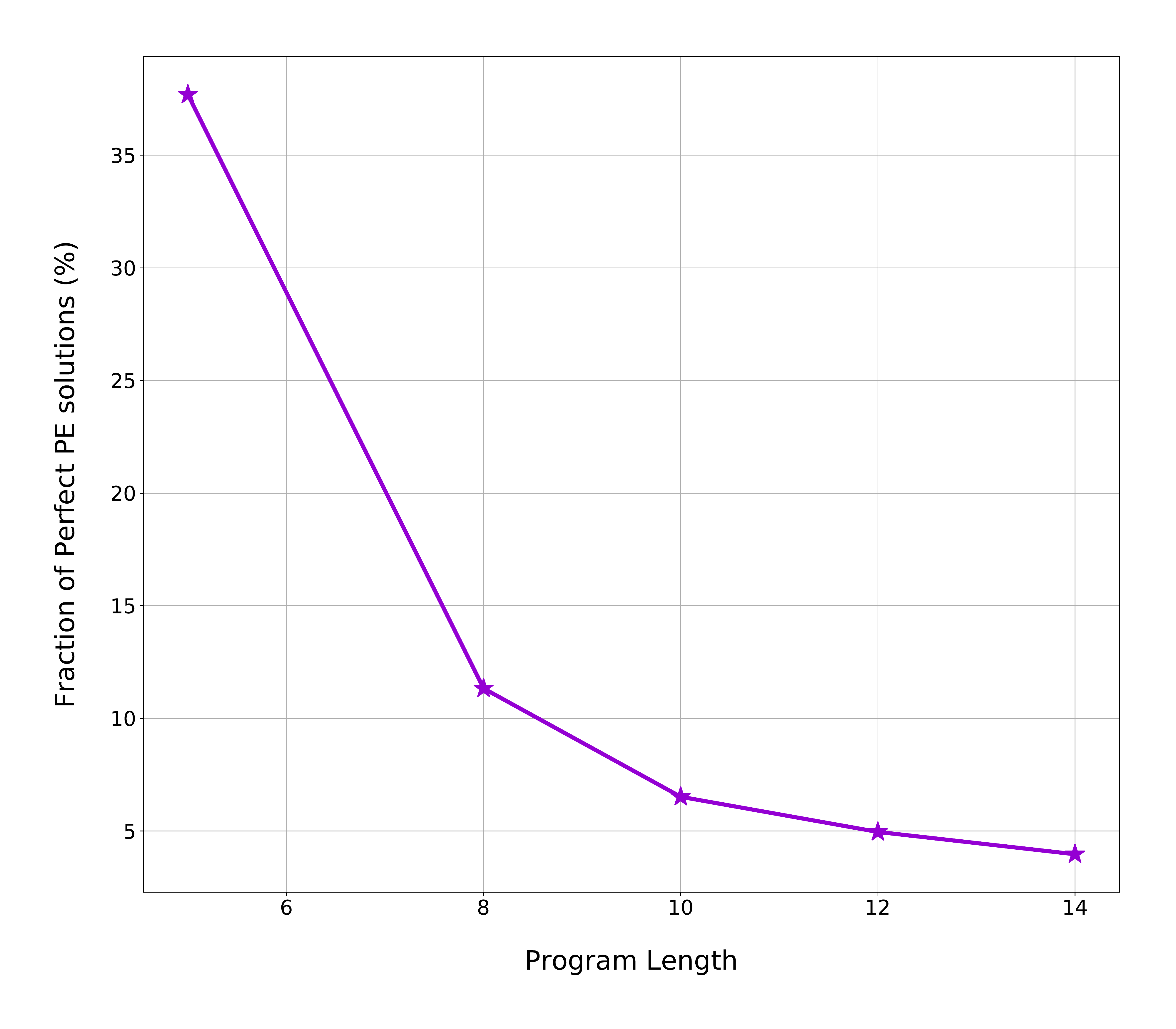}
  \caption{Fraction of perfect PE solutions with length of test programs for our best model for E2}
  \label{fig:perf_case}
  \vskip -0.15in
\end{figure}

One of the advantages of PEPS is that we may get a single PE solution which satisfies all IO examples (we call this perfect PE solution). In these cases, we do not even need to go to the CA and depending on when this perfect PE solution is discovered, it can lead to significant time savings (e.g., if the first PE solution discovered turns out to be a perfect solution, then the time taken to find the solution is equal to just the PEPS timeout which is upper bounded by 1/5th of the total timeout). Figure~\ref{fig:perf_case} shows the fraction of perfect PE solutions with the length of test programs for N-PEPS for E2. We see that as the program length increases, we have less chances to find a perfect PE solution. This is expected because it will be difficult for a single PE solution to satisfy all IO examples as the programs become lengthy (and hence complex). Note that even though we increase the depth of beam search based on the length of the test program, the overall budget (=5s) and the PEPS timeout (=0.8s in this case) remains the same across different lengths. This also means that for higher lengths, N-PEPS needs to rely more on CA to find a global solution.

\section{More insights into the workings of CA} 
\label{app:new_operator}
We tried to gain more insights into how our Cross Aggregator mechanism works. First, we looked into some general patterns learnt by the CA module. Second, we were interested in finding out how often does CA rely on copying from the PE solutions, how often does it generate new operators (in isolation with the GPS module) and how does it generate new operators. The last question is important as it helps us understand the generalization capabilities of CA outside the statements in the PE solutions. Even though we found it difficult to figure out a fixed scheme that worked across all the settings and different examples, by doing nearest-neighbour analysis, we were able to find some useful patterns that might shed some light towards answering this question.

\subsection{General Patterns Learnt by CA}
\label{app:gen_pattern_CA}

To look for general patterns, we inspected the representations of the final linear layer of our trained CA model (that is responsible for providing the logits used in the global statement prediction). The size of this weight matrix is $n_s \times Z$, where the $i$-th row can be interpreted as a learned embedding corresponding to the statement index $i$. We ran t-SNE on these embeddings and looked for interesting clusters. We found many cases where functionally similar statements or statements with similar signatures were clustered together. We give few examples of these patterns below:

\begin{itemize}
    \item[1.] \code{REVERSE b} almost overlaps with \code{SORT b}. This is interesting because both take in the list \code{b} and return another list without performing transformations on the elements in \code{b}.
    \item[2.] \code{MINIMUM b}, \code{MAXIMUM b}, \code{HEAD b}, and \code{TAIL b} are clustered together. This is interesting because all these operators select a single element from \code{b}.
    \item[3.] \code{FILTER (ODD) a} is close to \code{FILTER (ODD) b}. In this case, there is a difference of only the argument. For cases, where the prior statements in the program lead to transformations such that the contents of lists a and b are the same, like \code{b = SORT a} or \code{b = REVERSE a}, swapping \code{FILTER (ODD) a} with \code{FILTER(ODD) b} and vice-versa will give the same result.
\end{itemize}

\subsection{Overlap of PE Solutions with Global Solution}
\label{app:overlap_pe_global}
We wanted to see in how many cases do the operators present in the global solution also occur in one of the discovered PE solutions. This number gives us a rough estimate of how much can our attention mechanism do with just trying to copy these operators from one of the PE solutions when synthesizing the global solution. This is a rough estimate because we measure only the overlap of the operators and not statements, i.e., the arguments to the operators in the PE solutions and the global solution can be completely different. Specifically, we record the number of operators that overlap between the global solution (taken as the ground-truth program) and one of the PE solutions,  divided by the total number of statements in the ground truth programs across all cases. 

\begin{figure}[H]
  \centering
  \includegraphics[width=1.0\textwidth]{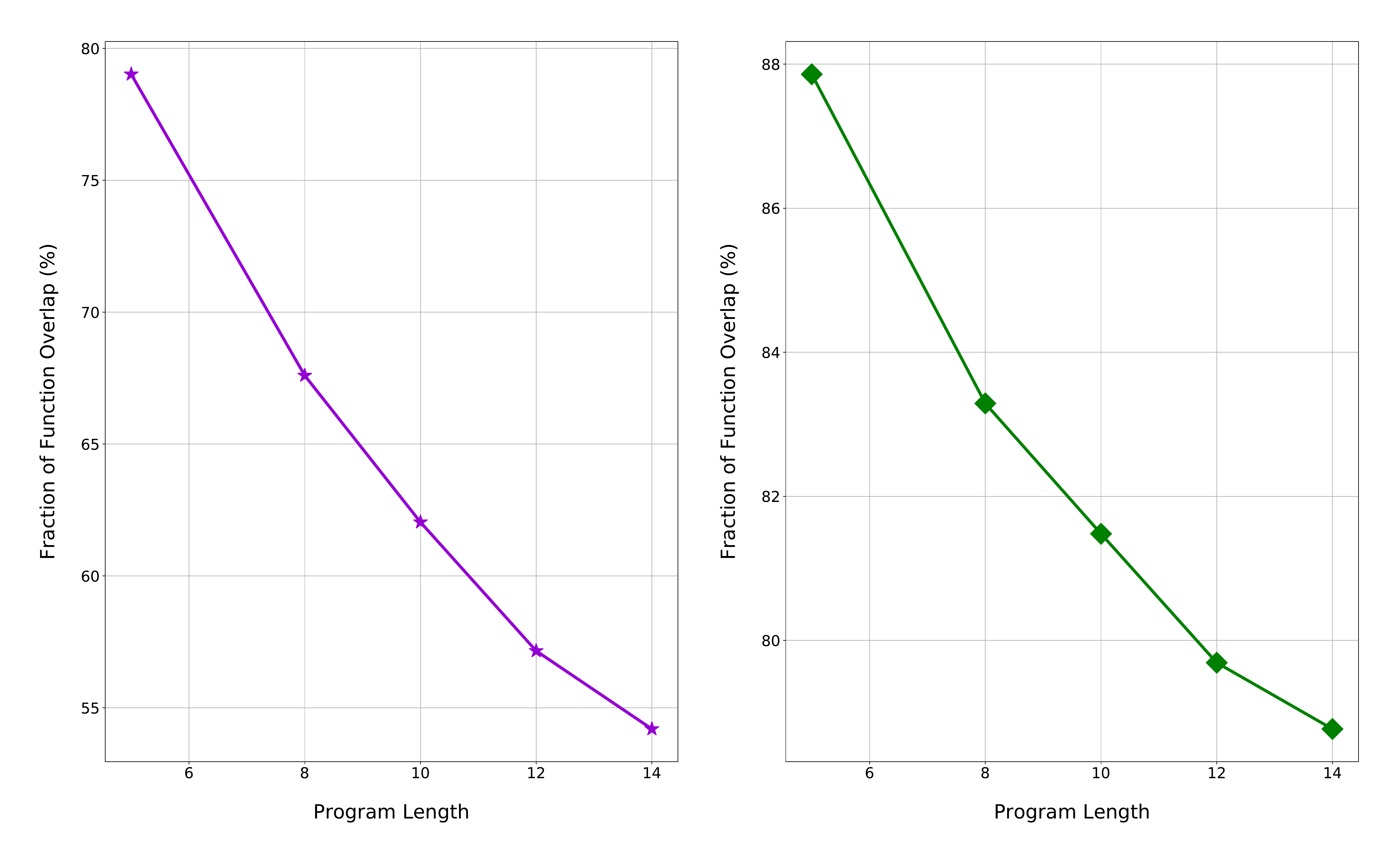}
  \caption{\textbf{Variation of fraction of operator overlap with length of test programs for our best model for E2}: \textbf{\textit{(Left)}} $\alpha < 1.0$ (CA + GPS); \textbf{\textit{(Right)}} $\alpha=1.0$ (CA alone)} 
  \label{fig:func_overlap}
  \vskip -0.15in
\end{figure}

The left part of Figure~\ref{fig:func_overlap} shows the variation of this number with different lengths of test programs for N-PEPS for E2 when $\alpha=0.8$ (which is the best-chosen hyperparameter value for this setting). We see that there is significant overlap between the operators indicating the quality of our PE solutions. However, the overlap decreases with length, which is also indicated by a decrease in overall success ratio with length (see left part of Figure 4). This is expected because we keep the same budget (PEPS timeout = 0.8s in this case) to discover PE solutions across all lengths. Improvements in the CA architecture focused to improve performance across longer length of programs in limited time budget, can be one of the potential directions to address this.

There is significant overlap (about 79\% for length 5), but not 100\% between the operators, highlighting that in many cases (21\% for length 5), N-PEPS performs discovery of new operators that are not present in the global solution. To further segregate the role of CA alone in the discovery of new operators as opposed to CA + GPS, we set $\alpha=1.0$ and analyzed the operator overlap. The right part of Figure~\ref{fig:func_overlap} shows this variation. As expected, the overlap percentages increase as compared to the case when $\alpha<1.0$. However, we can see that even when all the contribution to the global solution comes from the CA module alone, there is not a 100\% overlap between the operators and therefore, there are non-zero chances of discovery of new operators. From these plots, we can conclude that the CA is not merely a copy mechanism and is useful in scenarios where the discovered PE solutions are not significantly overlapping with the global solution.

\subsection{Sample cases where new operators are discovered}
\label{app:sample_new_gen}

\begin{figure}[H]
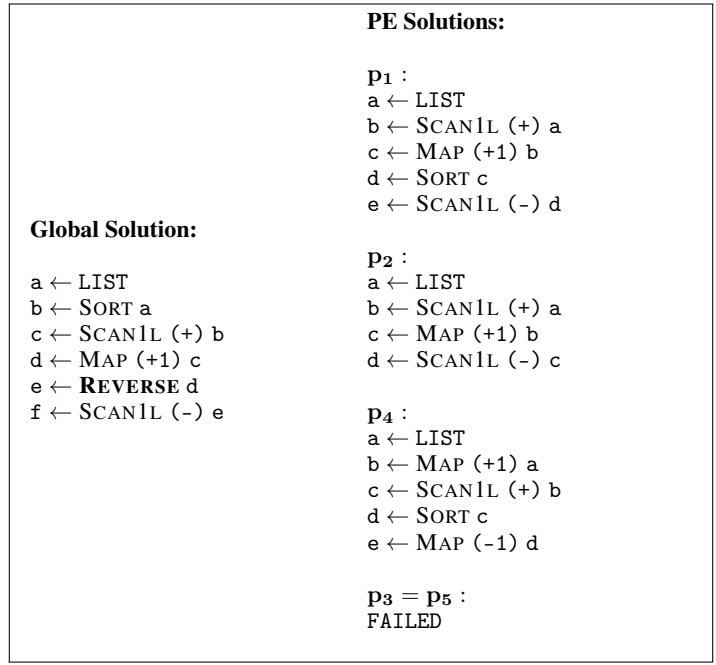

\centering
\fbox{
\scalebox{0.9}{
\begin{minipage}{.35\textwidth}
	\textbf{Global Solution:}\\ \\
    \texttt{a} $\gets$ \texttt{LIST}\\
    \texttt{b} $\gets$ \textsc{Sort} \texttt{a} \\
    \texttt{c} $\gets$ \textsc{Scan1l} \texttt{(+)} \texttt{b}\\
    \texttt{d} $\gets$ \textsc{Map} \texttt{(+1)} \texttt{c}\\
    \texttt{e} $\gets$ \textbf{\textsc{Reverse}} \texttt{d}\\
    \texttt{f} $\gets$ \textsc{Scan1l} \texttt{(-)} \texttt{e}\\
    \end{minipage}
    \hfill
    \begin{minipage}{.35\textwidth}
    \textbf{PE Solutions:}\\\\
    $\mathbf{p_1}:$\\
    \texttt{a} $\gets$ \texttt{LIST}\\
    \texttt{b} $\gets$ \textsc{Scan1l} \texttt{(+)} \texttt{a}\\
    \texttt{c} $\gets$ \textsc{Map} \texttt{(+1)} \texttt{b}\\
    \texttt{d} $\gets$ \textsc{Sort} \texttt{c} \\
    \texttt{e} $\gets$ \textsc{Scan1l} \texttt{(-)} \texttt{d}\\
 \\
    $\mathbf{p_2}:$\\
    \texttt{a} $\gets$ \texttt{LIST}\\
    \texttt{b} $\gets$ \textsc{Scan1l} \texttt{(+)} \texttt{a}\\
    \texttt{c} $\gets$ \textsc{Map} \texttt{(+1)} \texttt{b}\\
    \texttt{d} $\gets$ \textsc{Scan1l} \texttt{(-)} \texttt{c}\\
 \\
     $\mathbf{p_4}:$\\
    \texttt{a} $\gets$ \texttt{LIST}\\
    \texttt{b} $\gets$ \textsc{Map} \texttt{(+1)} \texttt{a}\\
    \texttt{c} $\gets$ \textsc{Scan1l} \texttt{(+)} \texttt{b}\\
    \texttt{d} $\gets$ \textsc{Sort} \texttt{c} \\
    \texttt{e} $\gets$ \textsc{Map} \texttt{(-1)} \texttt{d}\\
\\    
     $\mathbf{p_3 = p_5}:$\\
    \texttt{FAILED}\\
    \end{minipage}
}}
\caption{New operators discovered = 1, Total cases = 2169}
\label{fig:ex1}
\end{figure}

\begin{figure}[H]
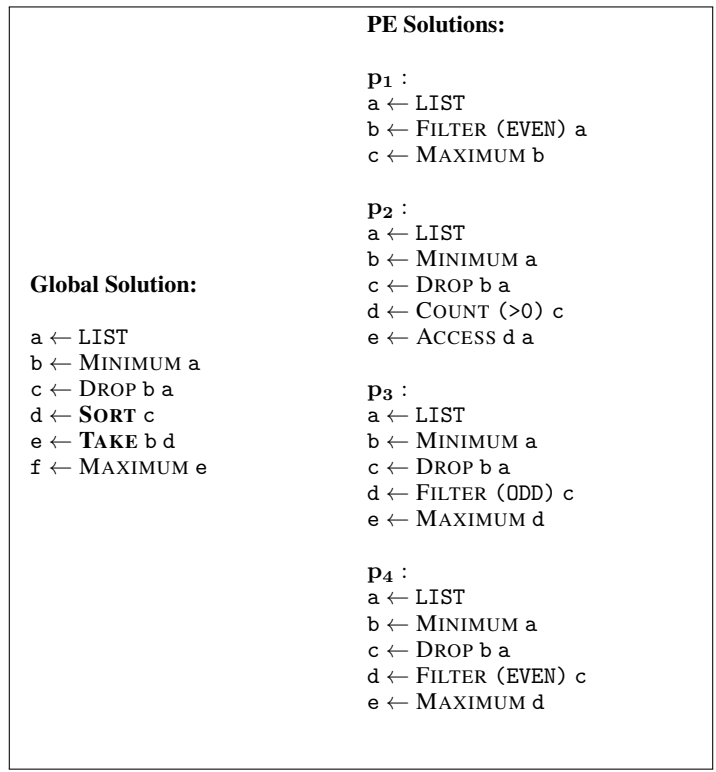

\centering
\fbox{
\scalebox{0.9}{
\begin{minipage}{.35\textwidth}
	\textbf{Global Solution:}\\ \\
    \texttt{a} $\gets$ \texttt{LIST}\\
    \texttt{b} $\gets$ \textsc{Minimum} \texttt{a} \\
    \texttt{c} $\gets$ \textsc{Drop} \texttt{b} \texttt{a}\\
    \texttt{d} $\gets$ \textbf{\textsc{Sort}} \texttt{c} \\
    \texttt{e} $\gets$ \textbf{\textsc{Take}} \texttt{b} \texttt{d}\\
    \texttt{f} $\gets$ \textsc{Maximum} \texttt{e} \\
    \end{minipage}
    
    \hfill
    
    \begin{minipage}{.35\textwidth}
    \textbf{PE Solutions:}\\\\
    $\mathbf{p_1}:$\\
    \texttt{a} $\gets$ \texttt{LIST}\\
    \texttt{b} $\gets$ \textsc{Filter} \texttt{(EVEN)} \texttt{a}\\
    \texttt{c} $\gets$ \textsc{Maximum} \texttt{b}\\
 \\
    $\mathbf{p_2}:$\\
    \texttt{a} $\gets$ \texttt{LIST}\\
    \texttt{b} $\gets$ \textsc{Minimum} \texttt{a}\\
    \texttt{c} $\gets$ \textsc{Drop} \texttt{b} \texttt{a}\\
    \texttt{d} $\gets$ \textsc{Count} \texttt{(>0)} \texttt{c}\\
    \texttt{e} $\gets$ \textsc{Access} \texttt{d} \texttt{a}\\
 \\
     $\mathbf{p_3}:$\\
    \texttt{a} $\gets$ \texttt{LIST}\\
    \texttt{b} $\gets$ \textsc{Minimum} \texttt{a} \\
    \texttt{c} $\gets$ \textsc{Drop} \texttt{b} \texttt{a}\\
    \texttt{d} $\gets$ \textsc{Filter} \texttt{(ODD)} \texttt{c}\\
    \texttt{e} $\gets$ \textsc{Maximum} \texttt{d} \\
\\    
     $\mathbf{p_4}:$\\
    \texttt{a} $\gets$ \texttt{LIST}\\
    \texttt{b} $\gets$ \textsc{Minimum} \texttt{a} \\
    \texttt{c} $\gets$ \textsc{Drop} \texttt{b} \texttt{a}\\
    \texttt{d} $\gets$ \textsc{Filter} \texttt{(EVEN)} \texttt{c}\\
    \texttt{e} $\gets$ \textsc{Maximum} \texttt{d} \\
    \\
    \end{minipage}
}}
\caption{New operators discovered = 2, Total cases = 1155}
\label{fig:ex2}
\end{figure}

Apart from the analysis done above, we wanted to gain further intuition of how the new operators are being discovered by the CA module. To this effect, we looked at sample cases of the generation of new operators by just CA (i.e., $\alpha=1.0$). In each box below (Figures~\ref{fig:ex1}-\ref{fig:ex5}), for test programs of length = 5, we report a case from the cases when the number of new operators discovered is 1, 2, 3, 4 and 5, respectively. The reported example shows the global solution discovered along with the corresponding PE solutions and is randomly chosen out of the total cases that fall within that category (i.e., not cherry-picked). For clarity, we bold the new operator in the global solution.

\begin{figure}[H]
\centering
\fbox{
\scalebox{0.9}{
\begin{minipage}{.35\textwidth}
	\textbf{Global Solution:}\\ \\
    \texttt{a} $\gets$ \texttt{LIST}\\
    \texttt{b} $\gets$ \textsc{Reverse} \texttt{a} \\
    \texttt{c} $\gets$ \textbf{\textsc{Count}} \texttt{(>0)} \texttt{b}\\
    \texttt{d} $\gets$ \textbf{\textsc{Access}} \texttt{c} \texttt{b} \\
    \texttt{e} $\gets$ \textbf{\textsc{Take}} \texttt{d} \texttt{b}\\
    \texttt{f} $\gets$ \textsc{Filter} \texttt{(ODD)} \texttt{e} \\
    \end{minipage}
    
    \hfill
    
    \begin{minipage}{.35\textwidth}
    \textbf{PE Solutions:}\\\\
    $\mathbf{p_1}:$\\
    \texttt{a} $\gets$ \texttt{LIST}\\
    \texttt{b} $\gets$ \textsc{Minimum} \texttt{a}\\
    \texttt{c} $\gets$ \textsc{Reverse} \texttt{a}\\
    \texttt{d} $\gets$ \textsc{Filter} \texttt{(ODD)} \texttt{c}\\
 \\
    $\mathbf{p_2 = p_3 = p_4}:$\\
    \texttt{a} $\gets$ \texttt{LIST}\\
    \texttt{b} $\gets$ \textsc{Filter} \texttt{(ODD)} \texttt{a}\\
    \texttt{c} $\gets$ \textsc{Reverse} \texttt{b}\\
   \\
     $\mathbf{p_5}:$\\
    \texttt{a} $\gets$ \texttt{LIST}\\
    \texttt{b} $\gets$ \textsc{Filter} \texttt{(<0)} \texttt{a}\\
   \\    
    \end{minipage}
}}
\caption{New operators discovered = 3, Total cases = 395}
\label{fig:ex3}
\end{figure}

\begin{figure}[H]
\centering
\fbox{
\scalebox{0.9}{
\begin{minipage}{.35\textwidth}
	\textbf{Global Solution:}\\ \\
    \texttt{a} $\gets$ \texttt{LIST}\\
    \texttt{b} $\gets$ \texttt{LIST}\\
    \texttt{c} $\gets$ \textsc{Tail} \texttt{b} \\
    \texttt{d} $\gets$ \textbf{\textsc{Count}} \texttt{(>0)} \texttt{a}\\
    \texttt{e} $\gets$ \textbf{\textsc{Drop}} \texttt{d} \texttt{b}\\
    \texttt{f} $\gets$ \textbf{\textsc{Sort}} \texttt{e} \\
    \texttt{g} $\gets$ \textbf{\textsc{Access}} \texttt{c} \texttt{f} \\
    \end{minipage}
    
    \hfill
    
    \begin{minipage}{.35\textwidth}
    \textbf{PE Solutions:}\\\\
    $\mathbf{p_1}:$\\
    \texttt{a} $\gets$ \texttt{LIST}\\
    \texttt{b} $\gets$ \texttt{LIST}\\
    \texttt{c} $\gets$ \textsc{Map} \textsc{(+1)} \texttt{b} \\
    \texttt{d} $\gets$ \textsc{Tail} \texttt{c}\\
 \\
    $\mathbf{p_2}:$\\
    \texttt{a} $\gets$ \texttt{LIST}\\
    \texttt{b} $\gets$ \texttt{LIST}\\
    \texttt{c} $\gets$ \textsc{Tail} \texttt{b}\\
\\    
    \end{minipage}
}}
\caption{New operators discovered = 4, Total cases = 119}
\label{fig:ex4}
\end{figure}

\begin{figure}[H]
\centering
\fbox{
\scalebox{0.9}{
\begin{minipage}{.35\textwidth}
	\textbf{Global Solution:}\\ \\
    \texttt{a} $\gets$ \texttt{LIST}\\
    \texttt{b} $\gets$ \texttt{LIST}\\
    \texttt{c} $\gets$ \textbf{\textsc{Sort}} \texttt{b} \\
    \texttt{d} $\gets$ \textbf{\textsc{Map}} \texttt{(+1)} \texttt{a}\\
    \texttt{e} $\gets$ \textbf{\textsc{Filter}} \texttt{(>0)} \texttt{d}\\
    \texttt{f} $\gets$ \textbf{\textsc{Reverse}} \texttt{c} \\
    \texttt{g} $\gets$ \textbf{\textsc{Zipwith}} \texttt{(+)} \texttt{f} \texttt{e}\\
    \end{minipage}
    
    \hfill
    
    \begin{minipage}{.35\textwidth}
    \textbf{PE Solutions:}\\\\
    $\mathbf{p_1 = p_2 = p_3 = p_4 = p_5}:$\\
    \texttt{FAILED}\\
    \end{minipage}
}}
\caption{New operators discovered = 5, Total cases = 15}
\label{fig:ex5}
\end{figure}

Looking at the above samples, there appears to be a trend where discovering fewer and shorter PE solutions leads to more new operators discovered. This may be attributed to the fact that when there is less information in the PE solutions, there is usually more of a need to generate new operators. The example in Figure~\ref{fig:ex5} is an extreme case of this, where no PE solutions were found, so all the operators need to be new.

\subsection{Nearest-neighbour analysis for new operators} 
\label{app:nn}
To gain intuition about how new operators are being generated by then CA module, we make two assumptions:
\begin{itemize}
    \item If a statement occurs frequently among the PE solutions, there is a high likelihood that it will also be present in the global solution. We find some evidence of this from our experiments in the paper where we show that the Sum-PEPS baseline performs better than GPS.
    \item If two statements $s_1$ and $s_2$ are close to each other in the output embedding space (with embeddings $e_1$ and $e_2$), they will also be similar in their corresponding logits. Here, we are assuming that $e_1 \approx e_2 \rightarrow x * e_1 \approx x*e_2$, with $x$ being the input activation.
\end{itemize}

With the above assumptions, for each of the examples provided in Appendix~\ref{app:sample_new_gen}, we calculated the top-10 nearest neighbours of the PE statements (using the representations obtained in a way described Appendix~\ref{app:gen_pattern_CA}). After this, we checked if the new operators in the global solution are present as part of the nearest neighbours of the PE statements. The presence of new operators points to a high likelihood of these being ranked higher in the beam search and hence being present in the global solution. In our analysis based on cases provided in Appendix~\ref{app:sample_new_gen}, we did observe this trend. We provide some instances below:

\begin{itemize}
    \item In Figure~\ref{fig:ex1} above, the statements containing the new operator \code{REVERSE} occur as the topmost neighbour (based on distance) of \code{SORT c , MAP (+1) b} , as well as among top-3 neighbours of \code{SCAN1L (+) a} . Note that the variation in certain cases from the general pattern observed before might be attributed to the two assumptions mentioned above not completely holding true in all cases.
    \begin{itemize}
        \item Top-3 neighbours of \code{SORT c} (occurs in $p_1, p_4$ ): [\code{REVERSE c, MAP (+1) c, COUNT (>0) c}]
        \item Top-3 neighbours of \code{MAP (+1) b} (occurs in $p_1, p_2$): [\code{REVERSE b, SORT b, COUNT (>0) b}]
        \item Top-3 neighbours of \code{SCAN1L (+1) a} (occurs in $p_1, p_2$): [\code{SCAN1L (-) a, SUM a, REVERSE a}]
    \end{itemize}
    
    \item In Figure~\ref{fig:ex2} above, new operator \code{SORT} is among the top-2 neighbours of \code{COUNT (>0) c}. Similarly, the new operator \code{TAKE} is among the top-2 neighbors of \code{DROP b a}.
    \begin{itemize}
        \item Top-3 neighbours of \code{COUNT (>0) c} (occurs in $p_2$): [\code{REVERSE c, SORT c, MAXIMUM c}]
        \item Top-5 neighbours of \code{DROP b a} (occurs in $p_2, p_3, p_4$): [\code{ACCESS b a, DROP b c, DROP b d, DROP c a, TAKE b a}]
    \end{itemize}

\end{itemize}




\end{document}